\pdfoutput=1 
\documentclass[10pt,twocolumn,letterpaper]{article}

\usepackage{cvpr}               
\usepackage{multirow} 
\usepackage{graphicx}
\usepackage[accsupp]{axessibility}  
\usepackage[pagebackref,breaklinks,colorlinks]{hyperref}
\usepackage{color}
\newcommand{\qs}[1]{\noindent\textcolor{red}{#1}}

\definecolor{cvprblue}{rgb}{0.21,0.49,0.74}

\def\paperID{3192} 
\def\confName{CVPR}
\def\confYear{2025}

\title{Dual-Interrelated Diffusion Model for Few-Shot Anomaly
Image Generation}

\author{
Ying Jin$^{1\ast}$,~~~Jinlong Peng$^{2}$\thanks{Equal contribution.} ,~~~Qingdong He$^{2\ast}$,~~~Teng Hu$^{3}$,~~~Jiafu Wu$^{2}$,~~~Hao Chen$^{1}$\\
Haoxuan Wang$^{1}$,~~~Wenbing Zhu$^{1}$,~~~Mingmin Chi$^{1}$\thanks{Corresponding author (This work was supported  by Natural Science Foundation of China under contract 62171139).},~~~Jun Liu$^{2}$,~~~Yabiao Wang$^{2,4\dagger}$\\
$^1$Fudan University,~~~$^2$Youtu Lab, Tencent,~~~$^3$Shanghai Jiao Tong University,~~~$^4$Zhejiang University\\
\small{$\{$yjin22, haochen22, hxwang23, wbzhu23$\}$@m.fudan.edu.cn, ~~mmchi@fudan.edu.cn}\\
\small{$\{$jeromepeng, yingcaihe, jiafwu,  juliusliu$\}$@tencent.com,~~hu-teng@sjtu.edu.cn, ~~yabiaowang@zju.edu.cn} \\
\qs{\url{https://github.com/yinyjin/DualAnoDiff}}
}

\begin{document}
\maketitle
\begin{abstract}
The performance of anomaly inspection in industrial manufacturing is constrained by the scarcity of anomaly data. To overcome this challenge, researchers have started employing anomaly generation approaches to augment the anomaly dataset.
However, existing anomaly generation methods suffer from limited diversity in the generated anomalies and struggle to achieve a seamless blending of this anomaly with the original image. Moreover, the generated mask is usually not aligned with the generated anomaly. In this paper, we overcome these challenges from a new perspective, simultaneously generating a pair of the overall image and the corresponding anomaly part.
We propose \textit{DualAnoDiff}, a novel diffusion-based few-shot anomaly image generation model, which can generate diverse and realistic anomaly images by using a dual-interrelated diffusion model, where one of them is employed to generate the whole image while the other one generates the anomaly part.
Moreover, we extract background and shape information to mitigate the distortion and blurriness phenomenon in few-shot image generation. 
Extensive experiments demonstrate the superiority of our proposed model over state-of-the-art methods in terms of diversity, realism and the accuracy of mask. Overall, our approach significantly improves the performance of downstream anomaly inspection tasks, including anomaly detection, anomaly localization, and anomaly classification tasks. Code will be made available.
\end{abstract}    
\section{Introduction}
\label{sec:intro}

\begin{figure}[!t]
\centering
\includegraphics[width=0.86\columnwidth]{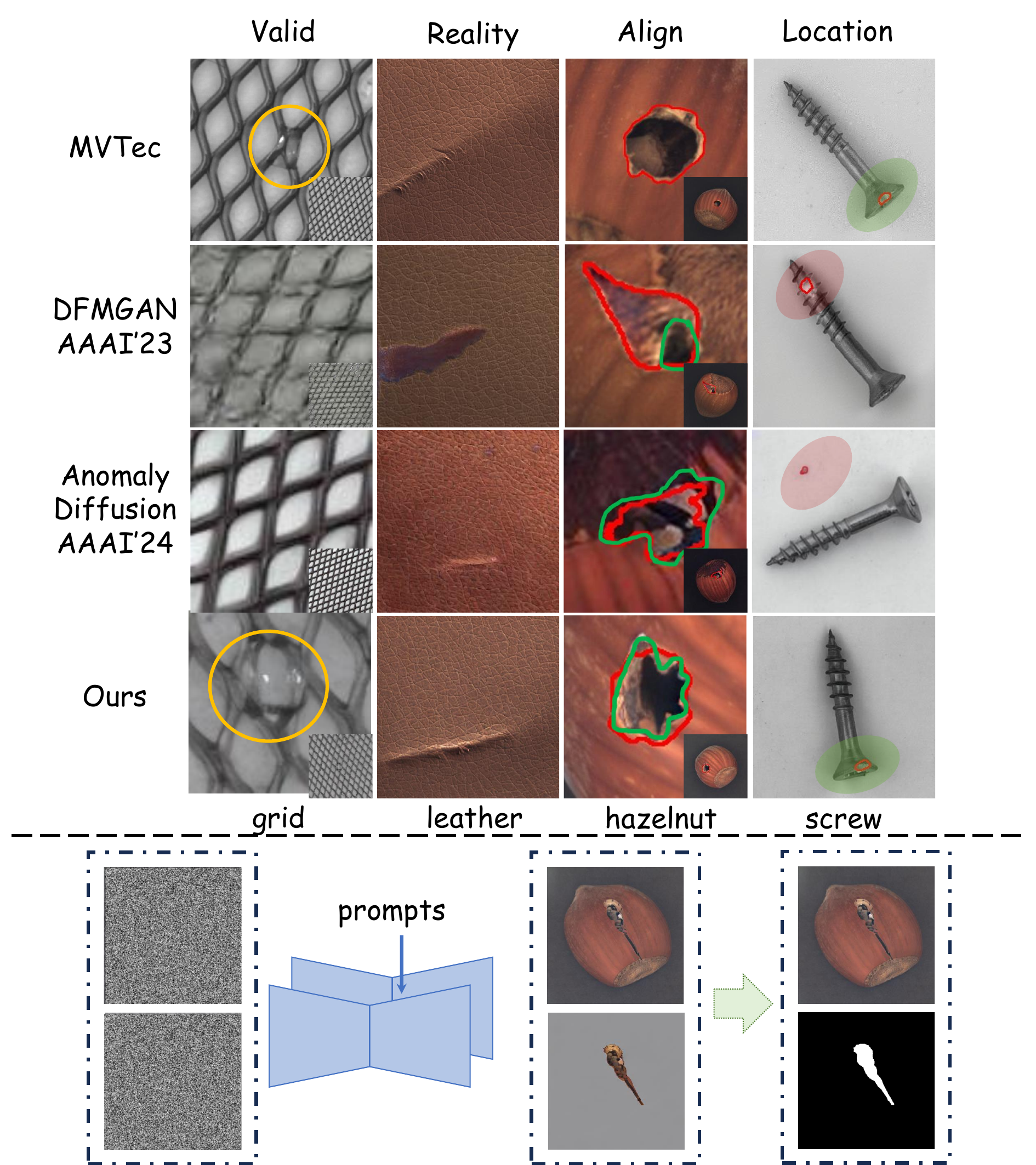} 
\caption{Top: Evaluating anomaly generation quality in four aspects: Whether generate valid anomaly, degree of realism, alignment of the mask, and whether the location of mask is reasonable, the results show that our generated results are better than the other methods. (Yellow area represents the valid generation, green indicates the correct mask or area, red indicates the generated mask or wrong area.) Bottom: Our model can simultaneously generate extensive anomaly image-mask pairs. }
\label{cmp}
\vspace{-0.4cm}
\end{figure}

Industrial anomaly inspection, i.e., anomaly detection, localization, and classification, plays an important role in industrial manufacture \cite{duan2023DFMGAN}. However, in real-world industrial production, anomaly samples are scarce. Therefore, the current mainstream anomaly inspection methods are either unsupervised methods which use only normal samples  \cite{roth2022patchcore,li2024musc} or semi-supervised method  \cite{zhang2023prn} which employ both the normal samples and a few anomaly data. Although these methods perform well in anomaly detection, they have limited performance in anomaly localization and can not deal with the task of anomaly classification  \cite{hu2024anomalydiffusion}. Therefore, researchers proposed anomaly generation methods to generate more anomaly data, to help achieve better performance by using supervised anomaly inspection.


Existing anomaly generation methods can be categorized into two groups, 1) \textit{model-free methods} randomly cut and paste patches from existing anomalies or anomalous texture datasets onto normal samples~\cite{lin2021croppaste,li2021cutpaste, zavrtanik2021draem}.
But the anomaly data synthesized by them are unrealistic.
2) \textit{Generative methods} employ generative models like GANs and diffusion models to generate anomaly data. Generally, GAN-based model  \cite{zhang2021defectgan,niu2020sdgan} requires a large amount of training data to achieve better generation performance and they can not generate masks.  DFMGAN \cite{duan2023DFMGAN} is firstly trained on normal data and then migrated to anomaly data to achieve a few-shot generation. This method also encounters the problem that the generated anomalies are not realistic enough \cite{peng2024frih} and the masks are not sufficiently aligned \cite{li2024tuning} because there is no explicit alignment constraint design. AnomalyDiffusion \cite{hu2024anomalydiffusion} based on the texture-inversion \cite{gal2022textinversion} technique of Diffusion \cite{rombach2022high}, separately learns the anomaly appearance and location information, then generates the anomaly on the masked normal samples. As AnomalyDiffusion only focuses on the part of anomaly, the generated anomalies do not blend realistically with the original image, and the masks generated individually may appear in the background of image.
To address these limitations, we propose DualAnoDiff, a novel few-shot anomaly image generation model that utilizes a dual-interrelated diffusion to simultaneously generate the overall image and the corresponding anomaly part. 
This approach can realize the effective integration of anomaly image and anomaly part, resulting in the generation of realistic and highly aligned anomaly image-mask data pairs with good diversity. Specifically, our model is built upon a pre-trained diffusion model and introduces two LoRA \cite{hu2021lora} to expand a single diffusion model into \textbf{dual-interrelated diffusion model}.
One branch, referred to as the global branch, is responsible for generating the overall anomaly image, while the anomalous branch generates the localized anomaly image. They exchange information through the \textbf{self-attention interaction module}.
This module combines the attention layers of global and anomalous branches and performs shared attention calculations, enabling the interaction and fusion of information in the dual-denoising process. This ensures the consistency between the generated overall anomaly image and the localized anomaly image.
Furthermore, to further preserve the invariance of the background, we introduce a \textbf{background compensation module} based on self-attention adaptive injection. 
This module involves adding noise to the background image, extracting the key and value from the intermediate feature layer, and applying adaptive fusion MLP to incorporate the background information into the global branch. 
It contributes to maintaining the accuracy of the background and the shape of the object in the generated images, while avoiding the coupling between the object and the background in the images.

Fig.\ref{cmp} shows our generated results outperform those of other methods in four key aspects. Moreover, extensive experiments have been conducted on MVTec AD \cite{bergmann2019mvtec} to quantitatively validate the superiority of the anomaly data generated by DualAnoDiff in downstream anomaly inspection tasks, and achieving a state-of-the-art level of performance in pixel-level anomaly detection with \textbf{99.1\% AUROC} and \textbf{84.5\% AP} score. 
%

\begin{figure*}[!ht]
\centering
\includegraphics[width=1.0\textwidth]{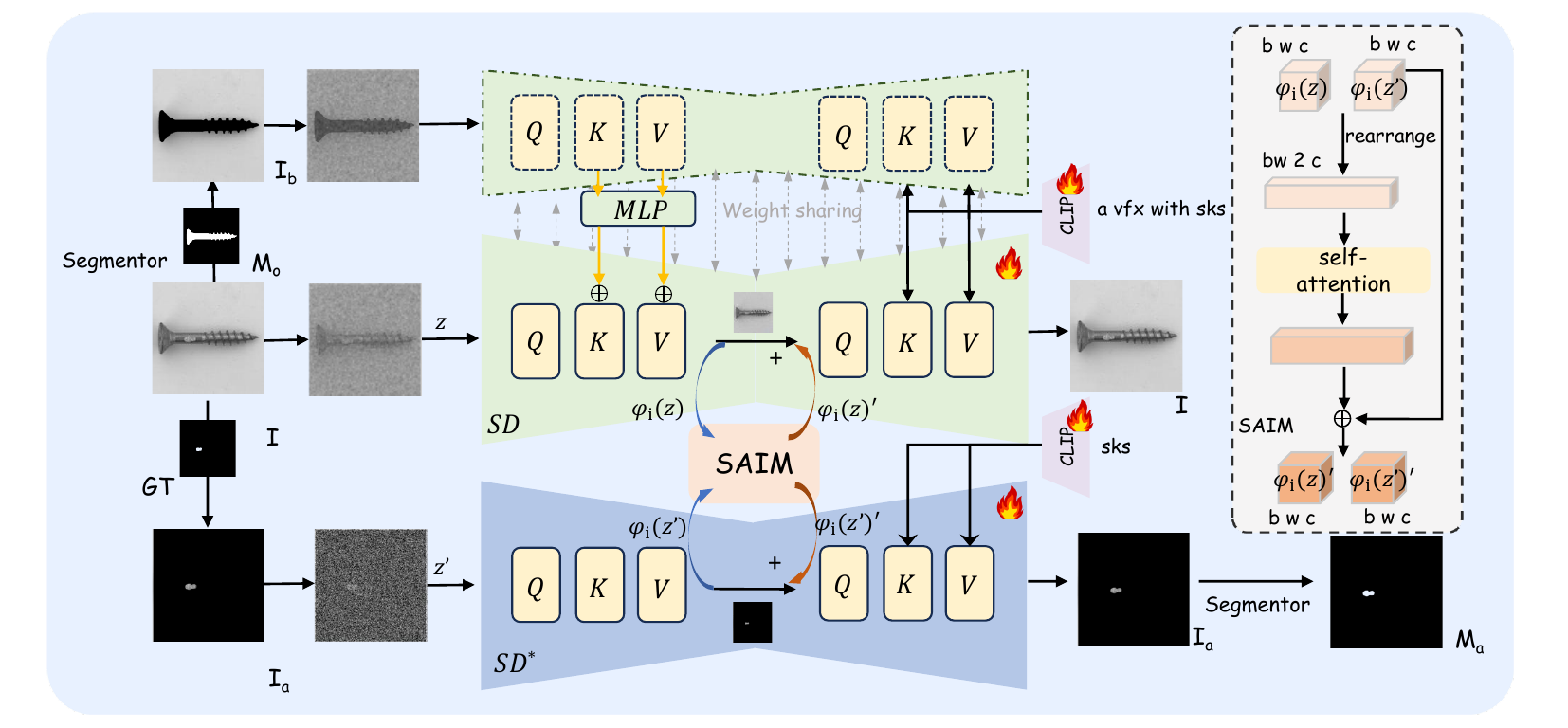}
\caption{The architecture of DualAnoDiff. 1) Two branches of DualAnoDiff generate the anomaly image and corresponding anomaly part simultaneously with different but nested prompts. 2) Two branches share the attention information after every attention block by Self-Attention Interaction Module (SAIM) during the denoising process to keep the consistency of generated images. 3) Background Compensation Module (BCM) extracts the Key, and Value of the background image and applies an adaptive fusion to SD, to help the model more focus on the object of the image.}
\label{pipeline}
\end{figure*}

Our contributions can be summarized as follows:
\begin{itemize}
    \item We propose DualAnoDiff, a novel few-shot diffusion-based anomaly generation method, which simultaneously generates both the overall image and the corresponding anomaly part with a highly aligned mask by a dual-interrelated diffusion model.
    \item We design a background compensation approach, which involves image backgrounds as control information and injects the intermediate feature into the denoising process of anomaly image, to enhance the stability and realism of the generated data.
    \item Extensive experiments demonstrate the superiority of our method over existing anomaly generation models in terms of both generation quality and performance of downstream anomaly inspection tasks.
\end{itemize}

\section{Related Work}
\label{sec:related_work}

\subsection{Few-shot Image Generation}
Few-shot image generation aims to generate new and diverse examples while preventing overfitting to the few training images \cite{zhu2024high, yi2024feditnet++}. It is highly susceptible to overfit with extremely limited training data (less than 10) and then generate highly similar images. 
FreezeD \cite{mo2020freeze} proposes modifying network weights, using various regularization techniques and data augmentation to prevent overfitting \cite{hu2024anomalydiffusion}. There are also methods \cite{hu2023phasic} to mitigate overfitting by pre-training on the source domain and subsequently migrating to fewer samples through cross-domain consistency losses to keep the generated distribution.
Textual Inversion \cite{gal2022textinversion} and Dreambooth \cite{ruiz2023dreambooth} encode a few images into the textual space of a pre-trained diffusion model to achieve diverse target customization generation which preserves its key visual features.
Although these methods can generate realistic images, they are incapable of generating pixel-level annotations which are essential for anomaly image generation tasks. In contrast, our method enables high-quality annotations to be readily obtained by simultaneously generating local anomaly images.
\subsection{Anomaly Inspection}
The anomaly inspection task consists of anomaly detection, localization, and classification \cite{duan2023DFMGAN}.
Due to the scarcity of abnormal data in industrial scenarios, most methods \cite{roth2022patchcore, gu2024rethinking, li2024musc,lee2024continuous,liu2023simplenet,wang2025softpatch+,wang2024pspu,wang2024real} use unsupervised methods and semi-supervised methods.
Reconstruction-based methods \cite{schlegl2019f, saa, he2024learning} detect anomalies by analyzing the residual image before and after reconstruction. 
Embedding-based methods \cite{lee2022cfa, cao2022informative, wang2023multimodal} utilize pre-trained networks to extract the image-level features and patch-level features, and then perform clustering according to the similarity between the features to detect the anomalies. All of those methods can only address the task of anomaly detection, while having limited performance in anomaly localization and being incapable of anomaly classification. Through the generation of abnormal images, these three tasks can be successfully accomplished, and our method achieves the state-of-the-art performance.
\subsection{Anomaly Generation}
Due to the scarcity of anomaly data, anomaly generation has emerged as a field of crucial significance.
DRAEM \cite{zavrtanik2021draem}, Cut-Paste \cite{li2021cutpaste}, Crop-Paste \cite{lin2021croppaste} and PRN \cite{zhang2023prn} crop and paste unrelated textures or existing anomalies into normal sample. These approaches can be somewhat effective, but the generated anomalies are completely unrealistic and there is no way to use them as anomaly classification tasks. Subsequently, GANs \cite{goodfellow2020generative} have been applied for anomaly generation due to their ability to generate high-fidelity images. SDGAN \cite{niu2020sdgan} and DefectGAN \cite{zhang2021defectgan} generate anomalies
on normal samples by learning from anomaly data. However, they require a large amount of anomaly data and cannot generate anomaly masks. DFMGAN \cite{duan2023DFMGAN} transfers a StyleGAN2 \cite{karras2020stylegan2} pretained on normal samples to the anomaly domain, but lacks generation realism and accurate alignment between generated anomalies and masks. Subsequently, Diffusion models have been more widely used due to their extreme generalizability. AnomalyDiffusion \cite{hu2024anomalydiffusion} learns the features of anomaly and the distribution of masks through text inversion \cite{gal2022textinversion} technique of diffusion, to generate the specified anomaly at the position of the corresponding mask of the normal image. However, Since the method learns the anomaly part and the mask separately, it makes the generated mask not necessarily located on the object, and the anomaly-object transition is not natural enough.
Nevertheless, through the utilization of a two-branch diffusion model along with a background supplement module, our methodology successfully accomplishes the full decoupling of diverse attributes within the anomaly image. As a result, the generated data are more realistic and diversified, and the performance of various downstream tasks is significantly enhanced.

\section{Method}
Given a set of a limited number of anomaly image-mask pairs, our goal is to learn the features of the anomaly images, and then generate more anomaly image-mask pairs that belong to the same item and anomaly type, while ensuring greater diversity in the distribution and appearance of the anomaly. 
In our method, we generate the overall image and the anomaly part, which is then segmented to get the corresponding mask.

\subsection{Preliminaries}
\textbf{Latent Diffusion Models.}
Stable Diffusion (SD), a variant of the latent diffusion model (LDM) \cite{rombach2022high}, serves as a text-guided diffusion model. To generate high-resolution images while enhancing computational efficiency in the training
process, it employs a pre-trained variational autoencoder (VAE) \cite{kingma2013vae} encoder $\mathcal{E}(\cdot)$ to map images into latent space and perform an iterative denoising process. Subsequently, the predicted images are mapped back into pixel space through the pre-trained VAE decoder $D(\cdot)$. $\epsilon_{\theta}$ is the denoising
network, for each denoising step, the simplified optimization objective is defined as follows:
\begin{equation}
L_{L D M}( \theta)= 
\mathbb{E}_{\mathcal{E}(x), \epsilon,t}
\left[\| \epsilon-\epsilon_{\theta}\left(z_{t}, t, \tau_{\theta}(c)\right) \|_{2}^{2}\right]
\end{equation}
where $\epsilon$ are latent noise, the text description $c$ is encoded by the CLIP \cite{radford2021learning} text encoder $\tau_{\theta}(\cdot)$ and then used to guide the diffusion denoising process.
\subsection{DualAnoDiff Framework}
We have been searching for methods that can generate both anomaly image and mask.  Inspired by Layerdiffusion \cite{zhang2024transparent}, we decompose an anomaly image into two parts, the overall anomaly image and the corresponding anomaly part, where the overall image refers to the whole anomaly image $I$, the anomaly part refers to the part that contains only the anomaly region $I_{a}$. $I_{a} = I \times M_{a}$ ($M_{a}$ is the mask of anomaly part).

As shown in Fig. \ref{pipeline}, the proposed \textit{DualAnoDiff} involves two interrelated diffusion models, and they share part information by the Self-attention Interaction Module. We freeze the weights of diffusion models and use two LORA \cite{hu2021lora} to fine-tune them. For ease of description, we denote the two diffusion models as $SD$ and $SD^\ast$, $SD$ denote the diffusion model to generate global image $I$, $SD^\ast$ denote the diffusion model to generate anomaly part $I_{a}$.

\noindent\textbf{Dual-Interrelated Diffusion.}
AnomalyDiffusion \cite{hu2024anomalydiffusion} primarily focuses on the anomaly part, which may result in generated anomaly images lacking a convincingly realistic appearance. However, generating the complete anomaly image poses challenges in obtaining the corresponding mask.
To address those limitations, our proposed model simultaneously generates both the overall image and the anomaly part. This novel approach overcomes challenges in generating realistic anomaly images while ensuring the availability of accurate masks.

First, we encode $I$, $I_{a}$ into latent space $z$ and $z'$, with $z = \varepsilon (I)$ and  $z'= \varepsilon (I_{a})$ by using VAE encoder $\varepsilon (\cdot)$. Next we employ a forward process to add noise into the latents with the same timestep $t$, and then learn to denoising during the backward process guided by different prompts.
Throughout these processes, information is shared and synchronized between the two diffusion models through the SAIM, enabling the model to effectively fit the training data pairs.
By generating the anomalies separately, this approach achieves two important objectives. 
Through a simple yet effective operation of adding two LoRA, it enhances the diversity and reality of the generated anomaly. Additionally, it ensures a highly aligned mask that accurately corresponds to the anomaly image.

\noindent\textbf{Nested Prompts.}
The goal of the dual diffusion is to generate an anomaly image and anomaly part pair ($I$ and $I_{a}$), which exhibits an inclusion relationship. To facilitate the model's understanding of the distinct entities within the image (the primary subject and the anomaly), we employ a pair of prompts designed to reflect this inclusion  relationship:
\vspace{-0.2cm}
\begin{align}
p: \medspace &a \medspace \mathbf x \medspace with \medspace \mathbf y  \notag\\
p': \medspace & \mathbf y
\end{align}
\vspace{-0.5cm}

where the prompts $p$ and $p'$ correspond to the anomaly image $I$ and the corresponding anomaly part $I_{a}$ respectively. Both prompts are encoded by the trainable text encoder $\tau_{\theta}(\cdot)$ and then injected into Unet \cite{ronneberger2015u} of Diffusion.

\begin{figure}[t]
    \centering
    \includegraphics[width=0.8\linewidth]{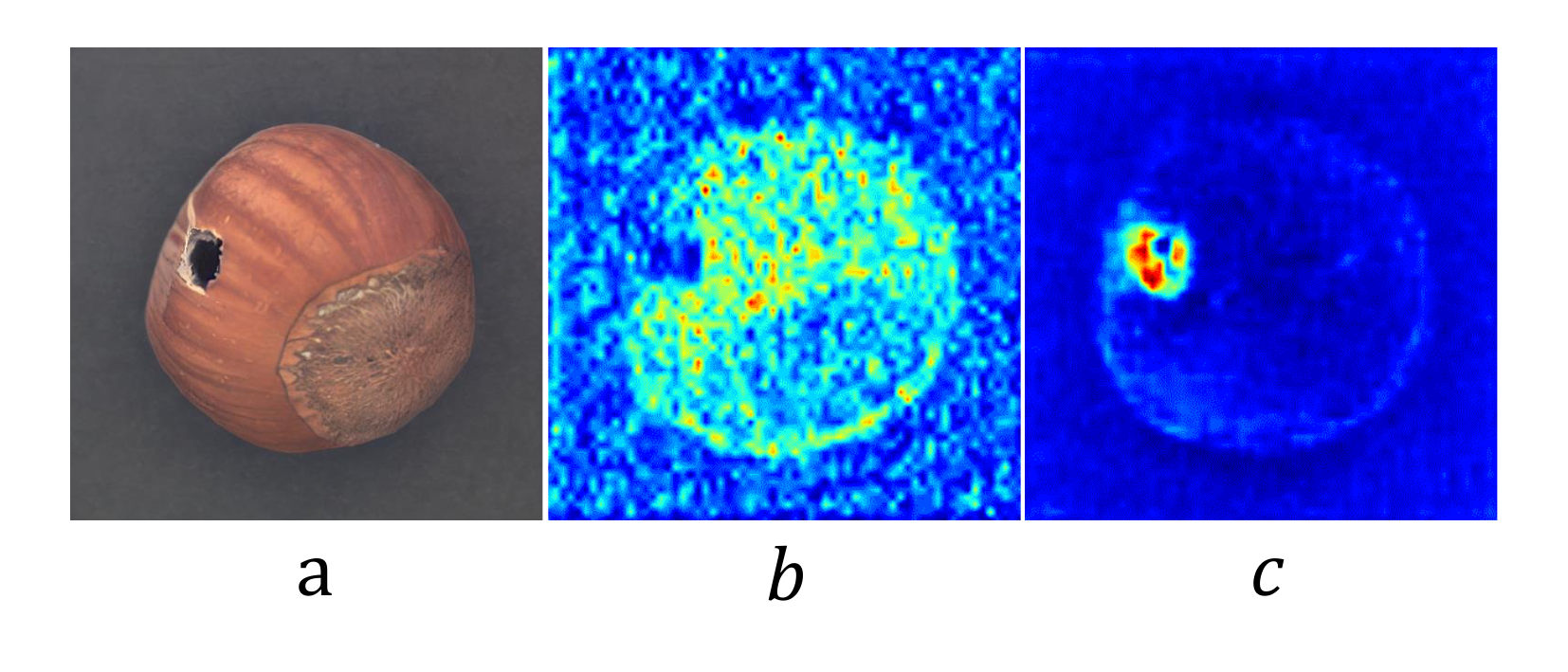}
    \vspace{-0.2cm}
    \caption{a is the image generated by SD, b and c are the cross attention maps of different text tokens in SD corresponding to the text of ``a vfx with'' and ``sks''.}
    \label{fig:vis}
\vspace{-0.5cm}
\end{figure}
The variables $x$ and $y$ can be the class name and anomaly name provided by the dataset. In our model, we use the $vfx$ and $sks$ which were suggested by DreamBooth \cite{ruiz2023dreambooth}. Those words have weak prior in both the language model and the diffusion model, making them easier to fit than other words, and can achieve better generation results, specially for high-prior words.
Fig.\ref{fig:vis} presents the generated result and visualizes the cross-attention maps between text token and vision. where $a$ is the anomaly image generated by $SD$. $b$ and $c$ are the 64 $\times$ 64 resolutions feature maps randomly extracted from the second half of the generation process in the Unet of $SD$. Where $b$ corresponds to the text ``$a$ $vfx$ $with$'', $c$ corresponds to ``$sks$''. It is evident that the model correctly separates the attributes of anomaly and object, and accurately associating them to the specified text as we want.
\noindent \textbf{Self-attention Interaction Module (SAIM).}
During training, $SD$ and $SD^\ast$ share the same timestep $t$ and  denoising simultaneously, they share information by SAIM after every attention blocks in the Unet of diffusion. For example, after the self-attention blocks is more likely to share the positional information and detailed information, the cross-attention blocks shares the semantic information.

In SAIM, we use attention to fuse the information from two branch, The shared step is formulated as:
\begin{align}
\varphi_{i}(\tilde{z}) = Rearrange&(Concat(\varphi_{i}(z),\varphi_{i}(z')))\notag\\
\varphi_{i}(\tilde{z})_{new} =& SelfAtt(\varphi_{i}(\tilde{z}))\\
\varphi_{i}(z)',\varphi_{i}(z')' = Split(&Rearrange(\varphi_{i}(\tilde{z})_{new}+\varphi_{i}(\tilde{z}))) \notag
\end{align}
where $\varphi_{i}$ is the intermediate representation of the Unet. The original shape of $\varphi_{i}(z)$ and $\varphi_{i}(z')$ are ``b w c'', where ``b'' denotes the batch size, ``w'' represents the spatial dimension, and ``c'' represents the channel dimension. We rearrange the $Concat(\varphi_{i}(z),\varphi_{i}(z'))$ to shape of ``bw 2 c'' to avoid displacement in space. $Concat$ and $Split$ are a set of corresponding operations that are used to aggregate and separate feature maps.

\begin{figure}[t]
    \centering
    \includegraphics[width=0.86\linewidth]{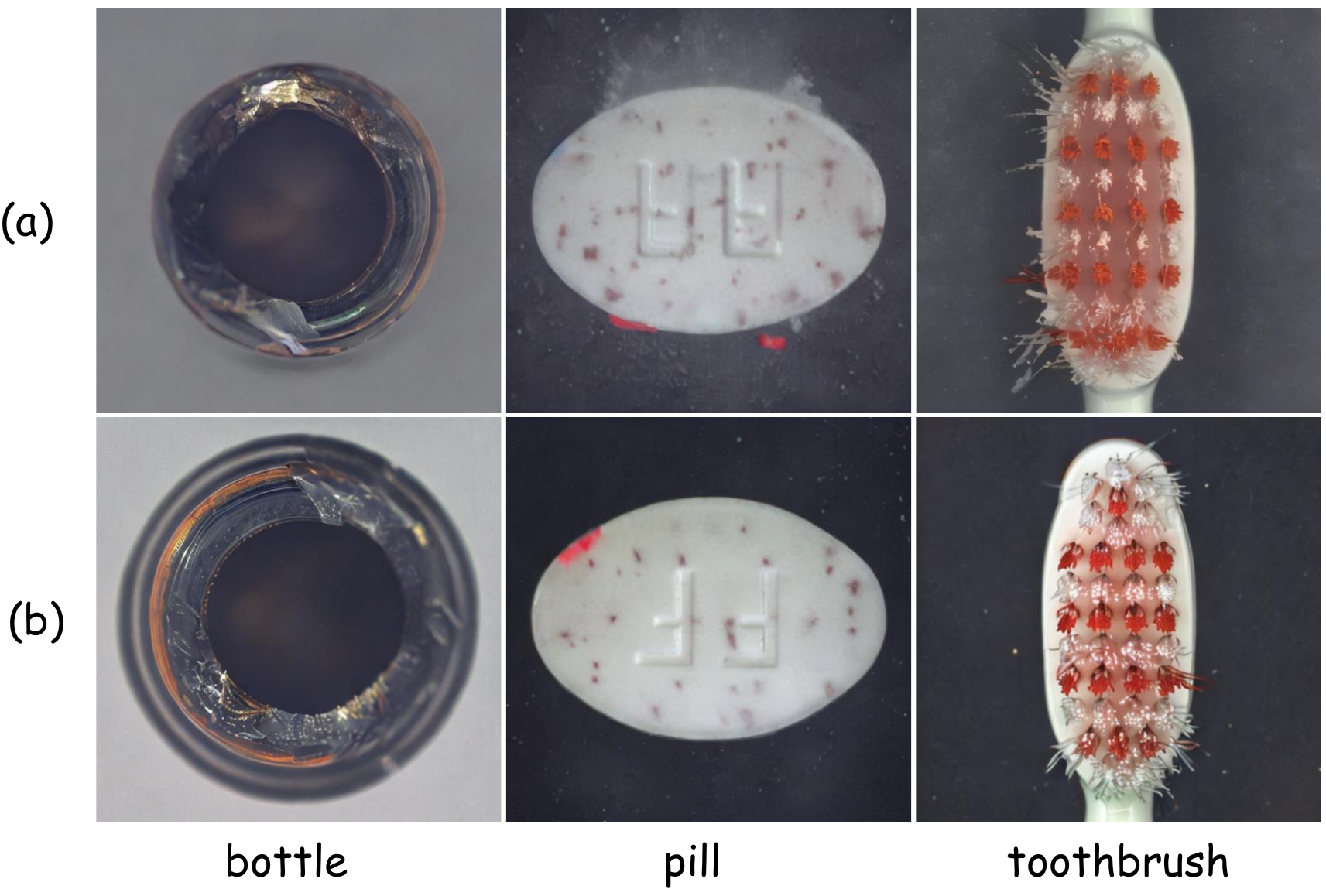}
    \caption{Comparison between the models without (a) and with (b) the Background Compensation Module.}
    \label{fig:background}
    \vspace{-0.4cm}
\end{figure}

\noindent\textbf{Loss Function.} With the two-stream structure of diffusion, our final training objective is expressed as follows:
\begin{align}
    \mathcal{L}=&\mathbb{E}_{\mathcal{E}(I), \epsilon,t}
    \left[\| \epsilon-\epsilon_{\theta}\left(z_{t}, t, \tau_{\theta}(p))\right) \|_{2}^{2}\right]\notag\\
    &+\mathbb{E}_{\mathcal{E}(I_{a}), \epsilon^\ast,t}
    \left[\| \epsilon^\ast-\epsilon_{\theta}^\ast\left(z'_{t}, t, \tau_{\theta}(p')\right) \|_{2}^{2}\right]
\end{align}
where, $\epsilon$ and $\epsilon^\ast$ are latent noise for the anomaly image and anomaly part, $t$ is the time step, $p$, $p'$ are the corresponding prompts, they are encoded by text encoder $\tau_{\theta}(\cdot)$ which is trainable. 

\noindent\textbf{Mask Generation.}
Thanks to our two-stream parallel structure, which generates the anomaly part as a single entity, obtaining the precise mask becomes straightforward. 
There are two ways to obtain a high-quality mask:
1). We utilize existing segmentation algorithms such as SAM \cite{kirillov2023segment}, U\textsuperscript{2}-Net \cite{qin2020u2} to obtain high-quality masks after generating the anomaly part image.
2) During the generation process, we can extract the average attention maps in $SD^\ast$ to compute the mask such as \cite{wang2023diffusion} which is widely used in semantic segmentation. 
In this paper, we use the first method.
\subsection{Background Compensation Module}
Although the model has achieved good results so far, there are still some problems in some cases, which is caused by the limited training data. 
Several bad cases of corresponding generated results are shown in Fig.\ref{fig:background} for categories bottle, pill, and toothbrush in the MVTec (the average training data for each category is 8).
The bottle was only partially generated. The edges of pill is blurred and the internal properties leakage into the background. Additionally, the toothbrush has two brush handles, which is an abnormal occurrence. Furthermore, all of those generated images lack sharpness and the background color does not maintain sufficient purity. 
In general, since the model is incapable of fully grasping the characteristics of those cases, there are problems such as objects being mixed with the background and objects being deformed.

To enhance the model's ability to learn the shape of object and focus more on object generation, we design a background compensation module. 
First, we employ U\textsuperscript{2}-Net to segment the image $I$ and obtain the object mask $M_{f}$, then get the background image $I_{b}=(1-M_{f})\times I$, $I_{b}$ will be processed by the $SD$ as same as $I$ (except for the SAIM). 
Then, we utilize $I_{b}$ as a condition, which contains both background and mask information. This allows us to to control the shape of the object while providing background context. 
We collect its features in self-attentions and inject them into the $K$,  $V$ of the $I$ in every corresponding self-attention steps.
The injection process can be formulated as:
\begin{align}
\varphi_{i}(z^{b}) =& SelfAtt(\varphi_{i}(z^{b})) \notag\\
Q =& W^{(i)}_{Q}\cdot\varphi_{i}(z)\notag\\
\varphi_{i}(z) =& \varphi_{i}(z) + \gamma MLP(\varphi_{i}(z^{b})) \notag\\
K = W^{(i)}_{K}\cdot&\varphi_{i}(z)  \quad V = W^{(i)}_{V}\cdot\varphi_{i}(z)
\end{align}
where $\gamma$ is a learnable scale factor initialized to be 0.1.
This design can help us to maximally preserve the generative effect of the mainstream SD, meanwhile use background information and the shape of the mask.
With the condition of $I_{b}$, the loss function becomes:
\begin{align}
    \mathcal{L}=&\mathbb{E}_{\mathcal{E}(I), \epsilon,t}\left[\left\|\epsilon-\epsilon_{\theta}\left(z_{t}, t, \tau_{\theta}(p),I_{b}\right)\right\|_{2}^{2}\right]\notag\\
    &+\mathbb{E}_{\mathcal{E}(I_{a}), \epsilon^\ast,t}
\left[\left\|\epsilon^\ast-\epsilon_{\theta}^\ast\left(z'_{t}, t, \tau_{\theta}(p')\right)\right\|_{2}^{2}\right]
\end{align}

\begin{table*}[t]
\centering
\small
\setlength{\tabcolsep}{1.6mm}
\begin{tabular}
{c|cc|cc|cc|cc|cc|cc|cc}
\toprule
\multirow{2}{*}{ Category } & \multicolumn{2}{c}{ CDC } & \multicolumn{2}{c}{ Crop-Paste } & \multicolumn{2}{c}{ SDGAN } & \multicolumn{2}{c}{ Defect-GAN } & \multicolumn{2}{c}{ DFMGAN } & \multicolumn{2}{c}{ AnomalyDiffusion } & \multicolumn{2}{c}{ Ours } \\
& IS  $\uparrow$  & IC-L  $\uparrow$  & IS  $\uparrow$  & IC-L  $\uparrow$  & IS  $\uparrow$  & IC-L  $\uparrow$  & IS  $\uparrow$  & IC-L & IS  $\uparrow$  & IC-L  $\uparrow$  & IS  $\uparrow$  & IC-L  $\uparrow$  & IS  $\uparrow$  & IC-L  $\uparrow$  \\
\midrule
bottle  & 1.52 & 0.04 & 1.43 & 0.04 & 1.57 & 0.06 & 1.39 & 0.07 & \underline{1.62} &  0.12  & 1.58 & \underline{0.19} &\textbf{ 2.17} &\textbf{0.43} \\
cable  & 1.97 & 0.19 & 1.74 & 0.25 & 1.89 & 0.19 & 1.70 & 0.22 & 1.96 &  0.25  &\underline{2.13}& \underline{0.41} & \textbf{2.15} &\textbf{0.43} \\
capsule & 1.37  & 0.06 & 1.23 & 0.05 & 1.49 & 0.03 & \underline{1.59} & 0.04 & \underline{1.59} &  0.11  & \underline{1.59} & \underline{0.21} & \textbf{1.62} &\textbf{0.32} \\
carpet & \underline{1.25}& 0.03& 1.17& 0.11& 1.18 & 0.11 & 1.24& 0.12& 1.23& 0.13& 1.16 & \underline{0.24} &\textbf{ 1.36}&\textbf{0.29} \\
grid& 1.97 & 0.07 & 2.00 & 0.12 & 1.95 & 0.10 & 2.01 & 0.12 & 1.97 & 0.13 & \underline{2.04} &\textbf{ 0.44} &\textbf{ 2.13} & \underline{0.42}\\
hazel\_nut & \underline{1.97} & 0.05 & 1.74 & 0.21 & 1.85 & 0.16 &  1.87  & 0.19 & 1.93 &  0.24  & \textbf{2.13} & \underline{0.31} & 1.94 &\textbf{0.35} \\
leather & 1.80 & 0.07 & 1.47 & 0.14 & 2.04 & 0.12 & \textbf{2.12} & 0.14 & \underline{2.06} &  \underline{0.17}  & 1.94 & \textbf{0.41} & 1.91 &\underline{0.35}\\
metal\_nut & 1.55 & 0.04 & \underline{1.56} & 0.15 & 1.45 & 0.28 & 1.47 & \underline{0.30} & 1.49 &  \textbf{0.32}  & \textbf{1.96 }& 0.30 & \underline{1.57} &\textbf{0.32} \\
pill & 1.56 & 0.06 & 1.49 & 0.11 & 1.61 & 0.07 & 1.61 & 0.10 & \underline{1.63} &  0.16  & 1.61 & \underline{0.26} &\textbf{1.82} &\textbf{0.38}\\
screw & 1.13 & 0.11 & 1.12 & 0.16 &  1.17  & 0.10 &  1.19  & 0.12 & 1.12 &  0.14  & \underline{1.28}  & \underline{0.30} & \textbf{1.43 }&\textbf{0.36}\\
tile & 2.10 & 0.12 & 1.83 &  0.20  & \underline{2.53} & 0.21 & 2.35 & 0.22 & 2.39 & 0.22 &\textbf{ 2.54} & \textbf{0.55 }& 2.40 &\underline{0.50} \\
toothbrush & 1.63 & 0.06 & 1.30 & 0.08 &  1.78  & 0.03 & \underline{1.85} &  0.03  & 1.82 &  0.18  & 1.68 & \underline{0.21} &\textbf{ 2.40 }&\textbf{0.48}\\
transistor & 1.61 & 0.13 & 1.39 & 0.15 & \textbf{1.76} & 0.13 & 1.47 & 0.13 & 1.64 & 0.25 & 1.57 &\textbf{ 0.34} & \underline{1.71} &\underline{0.33}\\
wood  & 2.05 & 0.03 & 1.95 & 0.23 & 2.12 & 0.25 & 2.19 & 0.29 &  2.12  &  0.35  & \textbf{2.33} & \underline{0.37} & \underline{2.24} &\textbf{0.40}\\
zipper  &  1.30  & 0.05 & 1.23 & 0.11 & 1.25 & 0.10 &  1.25  & 0.10 & 1.29 &  $\underline{0.27}$  & \underline{1.39} &  0.25 & \textbf{2.14} & \textbf{0.37}\\
\hline 
Average & 1.65 & 0.07 & 1.51 & 0.14 & 1.71 & 0.13 & 1.69 & 0.15 & 1.72 &  0.20  & \underline{1.80} & \underline{0.32} & \textbf{1.93} &\textbf{0.38}\\
\bottomrule
\end{tabular}
\caption{Comparison on IS and IC-LPIPS on MVTec dataset. Our model generates the most high-quality and diverse anomaly data, achieving the best IS and IC-LPIPS. Bold and underline represent optimal and sub-optimal results, respectively.}
\label{table1}
\vspace{-0.5cm}
\end{table*}

From another perspective, the dual-branch structure is equivalent to explicitly separating the two concepts of the object body and the anomaly in image. However, in the case of some images with backgrounds, two attributes are not enough. model may mix the object body with the solid-color background, thus resulting in the phenomenon shown in Fig. \ref{fig:background}(a). The BCM module further decouples the object body from the background by incorporating background information during training, thereby achieving a more stable generation effect in Fig. \ref{fig:background}(b).

\section{Experiments}

\subsection{Experiment Settings}
\noindent\textbf{Dataset.} We conduct experiments on MVTec \cite{bergmann2019mvtec} and follow the experimental setup of AnomalyDiffusion \cite{hu2024anomalydiffusion} using 1/3 of data for training, reserving the remaining 2/3 for testing.

\noindent\textbf{Implementation Details.} We train a model for each anomaly type separately and generate 1000 anomaly image-mask pairs for the downstream anomaly inspection tasks. More details of the experiment can be found in the supplementary materials.

\noindent\textbf{Metric.} \textbf{1) For generation}, we use Inception Score (IS) and Intra-cluster pairwise LPIPS distance (IC-LPIPS) \cite{ojha2021few} as AnomalyDiffusion \cite{hu2024anomalydiffusion} to measure the generation quality and generation diversity.  
\textbf{2) For anomaly inspection.} We use AUROC, Average Precision (AP), and the F1-max score to evaluate the accuracy of anomaly detection and localization, use Accuracy to evaluate the anomaly classification.
\subsection{Comparison in Anomaly Generation}
\noindent\textbf{Baseline.} 1) We choose CDC \cite{ojha2021few}, Crop-Paste\cite{lin2021croppaste}, SDGAN \cite{niu2020sdgan}, DefectGAN \cite{zhang2021defectgan}, DFMGAN \cite{duan2023DFMGAN} and AnomalyDiffusion \cite{hu2024anomalydiffusion} that can generate specific anomaly types to compare anomaly generation quality and classification.
2) We select DRAEM\cite{zavrtanik2021draem}, PRN \cite{zhang2023prn}, DFMGAN \cite{duan2023DFMGAN} and AnomalyDiffusion \cite{hu2024anomalydiffusion} which can generate anomaly image-mask pairs as comparative benchmarks to compare anomaly detection and localization.

\begin{table*}[t]
\small
\centering
\setlength{\tabcolsep}{0.4mm}
\begin{tabular}{c|cccc|cccc|cccc|cccc|cccc}
\toprule 
\multirow{2}{*}{ Category } & \multicolumn{4}{|c}{ DRAEM } & \multicolumn{4}{c}{ PRN } & \multicolumn{4}{c}{ DFMGAN } & \multicolumn{4}{c}{ AnomalyDiffusion } &\multicolumn{4}{c}{ Ours } \\
   & AUC-P & AP-P &  $F_{1}$-P & AP-I & AUC-P & AP-P &  $F_{1}$-P & AP-I &  AUC-P & AP-P &  $F_{1}$-P & AP-I & AUC-P & AP-P &  $F_{1}$-P & AP-I & AUC-P & AP-P &  $F_{1}$-P & AP-I \\
\midrule
bottle & 96.7 & 80.2 & 74.0 & 99.8 & 97.5 & 76.4 & 71.3 & 98.4 & 98.9 & 90.2 & 83.9 & 99.8 & \underline{99.4} & \textbf{94.1} & \textbf{87.3} & \underline{99.9} & \textbf{99.5} & \underline{93.4} & \underline{85.7} & \textbf{100} \\
cable & 80.3 & 21.8 & 28.3 & 83.2 & 94.5 & 64.4 & 61.0 & 92.0 & 97.2 & 81.0 & 75.4 & 97.8 & \textbf{99.2} & \textbf{90.8} & \textbf{83.5} & \textbf{100} & \underline{97.5} & \underline{82.6} & \underline{76.9} & \underline{98.3} \\
capsule & 76.2 & 25.5 & 32.1 & 98.7 & 95.6 & 45.7 & 47.9 & 95.8 & 79.2 & 26.0 & 35.0 & 98.5 & \underline{98.8} & \underline{57.2} & \underline{59.8} & \textbf{99.9} & \textbf{99.5} & \textbf{73.2} & \textbf{67.0} & \underline{99.2} \\
carpet & 92.6 & 43.0 & 41.9 & 98.7 & 96.4 & 69.6 & 65.6 & 97.8 & 90.6 & 33.4 & 38.1 & 98.5 & \underline{98.6} & \underline{81.2} & \underline{74.6} & \underline{98.8} & \textbf{99.4} & \textbf{89.1} & \textbf{80.2} & \textbf{99.9} \\
grid & \textbf{99.1} & \textbf{59.3} & \underline{58.7} & \textbf{99.9} & \underline{98.9} & \underline{58.6} & \textbf{58.9} & 98.9 & 75.2 & 14.3 & 20.5 & 90.4 & 98.3 & 52.9 & 54.6 & 99.5 & 98.5 & 57.2 & 54.9 & \underline{99.7} \\
hazelnut & 98.8 & 73.6 & 68.5 & \textbf{100} & 98.0 & 73.9 & 68.2 & 96.0 & \underline{99.7} & 95.2 & 89.5 & \textbf{100} & \textbf{99.8} & \underline{96.5} & \underline{90.6} & \underline{99.9} & \textbf{99.8} & \textbf{97.7} & \textbf{92.8} & \textbf{100} \\
leather & 98.5 & 67.6 & 65.0 & \textbf{100} & 99.4 & 58.1 & 54.0 & \underline{99.7} & 98.5 & 68.7 & 66.7 & \textbf{100} & \underline{99.8} & \underline{79.6} & \underline{71.0} & \textbf{100} & \textbf{99.9} & \textbf{88.8} & \textbf{78.8} & \textbf{100} \\
metal\_nut & 96.9 & 84.2 & 74.5 & 99.6 & 97.9 & 93.0 & 87.1 & 99.5 & 99.3 & \underline{98.1} & \textbf{94.5} & 99.8 & \textbf{99.8} & \textbf{98.7} & \underline{94.0} & \textbf{100} & \underline{99.6} & 98.0 & 93.0 & \underline{99.9} \\
pill & 95.8 & 45.3 & 53.0 & 98.9 & 98.3 & 55.5 & 72.6 & 97.8 & 81.2 & 67.8 & 72.6 & 91.7 & \textbf{99.8} & \textbf{97.0} & \textbf{90.8} & \textbf{99.6} & \underline{99.6} & \underline{95.8} & \underline{89.2} & \underline{99.0} \\
screw & 91.0 & 30.1 & 35.7 & \underline{96.3} & 94.0 & 47.7 & 49.8 & 94.7 & 58.8 & 2.2 & 5.3 & 64.7 & \underline{97.0} & \underline{51.8} & \underline{50.9} & \textbf{97.9} & \textbf{98.1} & \textbf{57.1} & \textbf{56.1} & 95.0 \\
tile & 98.5 & 93.2 & 87.8 & \textbf{100} & 98.5 & 91.8 & 84.4 & \underline{96.9} & \underline{99.5} & \textbf{97.1} & \textbf{91.6} & \textbf{100} & 99.2 & \underline{93.9} & 86.2 & \textbf{100} & \textbf{99.7} & \textbf{97.1} & \underline{91.0} & \textbf{100} \\
toothbrush & 93.8 & 29.5 & 28.4 & \underline{99.8} & 96.1 & 46.4 & 46.2 & \textbf{100} & 96.4 & \underline{75.9} & \underline{72.6} & \textbf{100} & \textbf{99.2} & \textbf{76.5} & \textbf{73.4} & \textbf{100} & \underline{98.2} & 68.3 & 68.6 & 99.7 \\
transistor & 76.5 & 31.7 & 24.2 & 80.5 & 94.9 & 68.6 & 68.4 & 88.9 & 96.2 & 81.2 & 77.0 & 92.5 & \textbf{99.3} & \textbf{92.6} & \textbf{85.7} & \textbf{100} & \underline{98.0} & \underline{86.7} & \underline{79.6} & \underline{93.7} \\
wood & 98.8 & \underline{87.8} & \underline{80.9} & \textbf{100} & 96.2 & 74.2 & 67.4 & 92.7 & 95.3 & 70.7 & 65.8 & 99.4 & \underline{98.9} & 84.6 & 74.5 & 99.4 & \textbf{99.4} & \textbf{91.6} & \textbf{83.8} & \underline{99.9} \\
zipper & 93.4 & 65.4 & 64.7 & \textbf{100} & 98.4 & 79.0 & 73.7 & 99.7 & 92.9 & 65.6 & 64.9 & \underline{99.9} & \underline{99.4} & \underline{86.0} & \underline{79.2} & \textbf{100} & \textbf{99.6} & \textbf{90.7} & \textbf{82.7} & \textbf{100} \\
\hline
Average & 92.2 & 54.1 & 53.1 & 97.0 & \underline{96.9} & 66.2 & 64.7 & 96.6 & 90.0 & 62.7 & 62.1 & 94.8 & \textbf{99.1} & \underline{81.4} & \underline{76.3} & \textbf{99.7} & \textbf{99.1} & \textbf{84.5} & \textbf{78.8} & \underline{98.9} \\
\bottomrule
\end{tabular}
\vspace{-0.3cm}
\caption{\textbf{Quantitative result for anomaly detection.} Comparison on pixel-level anomaly localization and image-level anomaly detection on MVTec dataset by training an U-Net on the generated data from DRAEM, PRN, DFMGAN, AnomalyDiffusion and our model.}
\label{localization}
\vspace{-0.4cm}
\end{table*}

\begin{figure}[!t]
\centering
\includegraphics[width=1\linewidth]{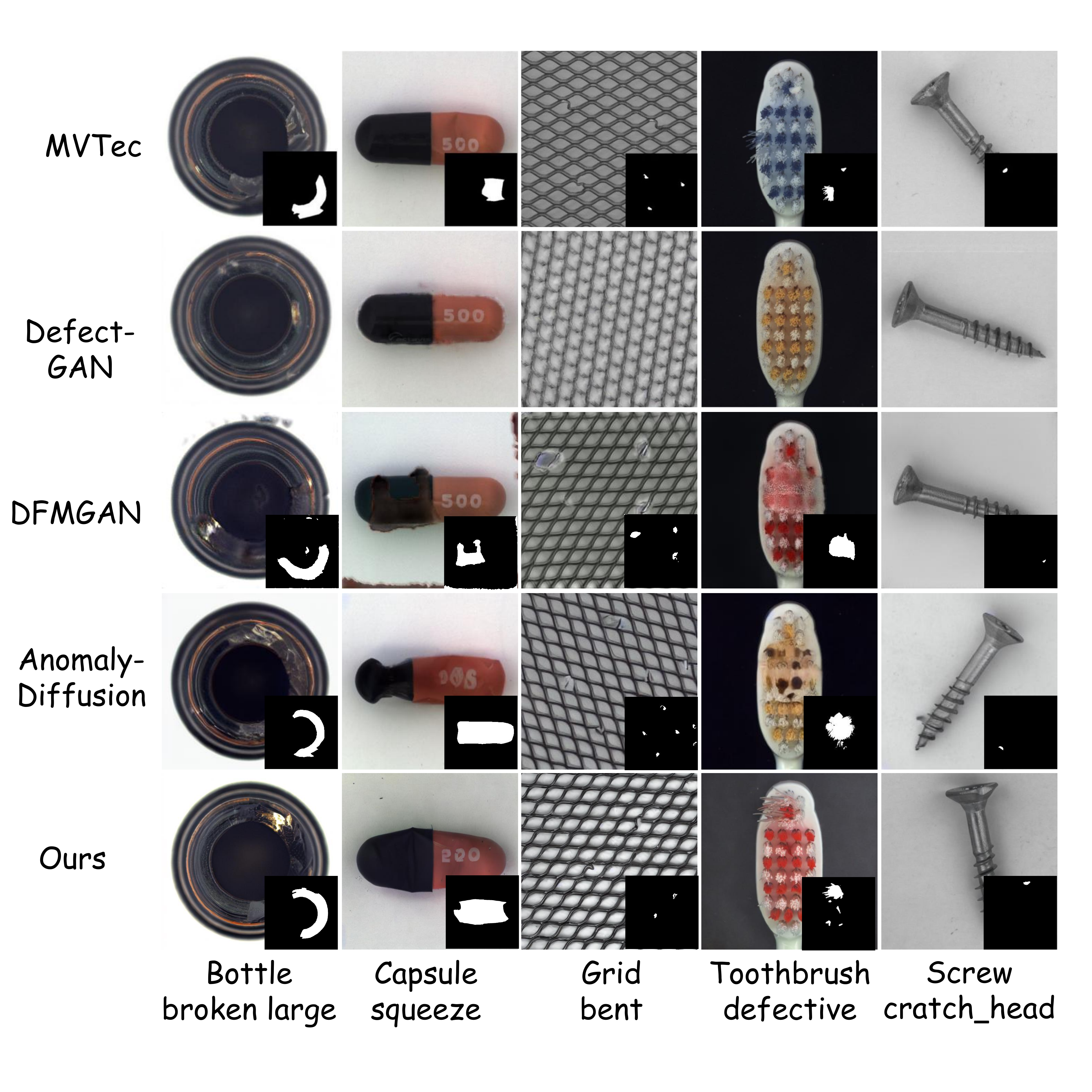}
\vspace{-0.7cm}
\caption{Comparison of the generation results on MVTec. Our model excels in generating high-quality anomaly images that are accurately aligned with the anomaly masks.}
\label{fig:generations}
\vspace{-0.2cm}
\end{figure}

\begin{table}[!t]
\setlength{\tabcolsep}{1mm}
\small
\begin{tabular}{c|cccccc}
\toprule 
Category & DiffAug  & Crop-Paste & DFMGAN & AnoDiff & Ours \\
\midrule 
bottle & 48.84 & 52.71  &  56.59  &  \textbf{90.70} & \underline{79.07} \\
cable & 21.36 & 32.81 &  45.31  &  \underline{67.19}  & \textbf{78.12}\\
capsule & 34.67 & 32.8 &  37.23  &  \underline{66.67} &\textbf{70.67}  \\
carpet & 35.48 & 27.96 &  47.31  &  \underline{58.06} &\textbf{79.03} \\
grid & 28.33  & 28.33  &  40.83  &  \underline{42.50}  &\textbf{80.0}\\
hazelnut & 65.28  & 59.03 &  81.94  &  \underline{85.42}  &\textbf{89.58}\\
leather & 40.74 & 34.39  &  49.73  &  \underline{61.90}  & \textbf{90.48}\\
metalnut & 58.85 &  59.89  &  \underline{64.58}  & 59.38 & \textbf{89.06}\\
pill &  29.86 & 26.74  & 29.52 & \textbf{59.38}  & \underline{56.25}\\
screw & 25.10 & 28.81  &  37.45  &  \underline{48.15} &\textbf{70.37}\\
tile & 59.65 & 68.42  &  74.85  &  \underline{84.21} & \textbf{100}\\
transistor & 38.09  & 41.67  &  52.38  &  \underline{60.71} & \textbf{71.43}  \\
wood & 41.27  & 47.62  &  49.21  &  \underline{71.43} &\textbf{85.71}\\
zipper & 22.76  & 26.42 &  \underline{27.64}  &  \textbf{69.51}  &\textbf{75.61}\\
\hline 
Average & 39.31 & 40.55 &  49.61  &  \underline{66.09} & \textbf{79.67}  \\
\bottomrule
\end{tabular}
\vspace{-0.2cm}
\caption{\textbf{Quantitative result for anomaly classification.} Comparison on the classification accuracy(\%) trained on the generated data by the anomaly generation models with a ResNet-18. The higher classification accuracy indicates that the generated data is more consistent with the distribution of real data.}
\label{classification}
\vspace{-0.8cm}
\end{table}

\noindent\textbf{Anomaly Generation Quality.}
Tab. \ref{table1} shows the results of the quantification metrics for image quality.
For each anomaly category, we allocate 1/3 of the anomaly data for training and generate 1000 anomaly images to compute IS and IC-LPIPS.
Through this process, it is clearly demonstrated that our method achieves the best results in both the IS and IC-LPIPS metrics.  Moreover, it further shows that our model generates anomaly data with both the highest quality and diversity.

\begin{table*}[t]
\small
\centering
\setlength{\tabcolsep}{1.2mm}
\begin{tabular}{c|cccccc|cccc}
\toprule
\multirow{2}{*}{ Category } & \multicolumn{6}{c}{ Unsupervised } & \multicolumn{4}{c}{ Supervised } \\
  & DRAEM & SSPCAB & CFA & RD4AD & PatchCore & Musc & DevNet & DRA & PRN & Ours \\
\midrule 
bottle &  99.1/88.5  & 98.9/88.6 &  98.9/50.9  &  98.8/51.0  &  97.6/75.0 & 98.5/82.8 &  96.7/67.9  &  91.7/41.5  &  99.4/92.3  & \textbf{99.5}/\textbf{93.4} \\
cable  & 94.8/61.4 &  93.1/52.1  &  98.4/79.8  & \textbf{98.8}/77.0 & 96.8/65.9 & 96.2/58.8  & 97.9/67.6 &  86.1/34.8  & \textbf{98.8}/78.9 &  97.5/\textbf{82.6}  \\
capsule &97.6/47.9 & 90.4/48.7 & 98.9/71.1& 99.0/60.5&  98.6/46.6 & 98.9/52.7 & 91.1/46.6& 88.5/11.0 & 98.5/62.2 &  \textbf{99.5}/\textbf{73.2} \\
 carpet  &  96.3/62.5  &  92.3/49.1  &  99.1/47.7  &  \textbf{99.4}/46.0  &  98.7/65.0 & \textbf{99.4}/75.3 &  94.6/19.6  &  98.2/54.0  & 99.0/82.0 &  \textbf{99.4}/\textbf{89.1}  \\
 grid   &  99.5/53.2  & \textbf{99.6}/58.2 & 98.6/\textbf{82.9} &  98.0/75.4  &  97.2/23.6  &98.6/37.0 & 90.2/44.9  & 86.2/28.6 &  98.4/45.7  &  98.5/57.2  \\
 hazelnut   & 99.5/88.1 & 99.6/94.5 &  98.5/80.2  & 94.2/57.2 &  97.6/55.2 & 99.3/74.4 & 76.9/46.8 & 88.8/20.3 & \textbf{99.7}/93.8 & \textbf{99.8}/\textbf{97.7} \\
 leather  &  98.8/68.5  &  97.2/60.3  & 96.2/60.9 &  96.6/53.5  & 98.9/43.4 &99.7/62.6 & 94.3/66.2  & 97.2/ 5.1 &  99.7/69.7  & \textbf{99.9}/\textbf{88.8} \\
 metal\_nut & 98.7/91.6 &99.3/95.1& 98.6/74.6& 97.3/53.8 & 97.5/86.8 & 87.5/49.6 & 93.3/57.4 & 80.3/30.6 & \textbf{99.7}/\textbf{98.0} &  99.6/\textbf{98.0} \\ 
pill   &  97.7/44.8  &  96.5/48.1  & 98.8/67.9 &  98.4/58.1  & 97.0/75.9 & 97.6/65.6 & 98.9/79.9 &  79.6/22.1  &  99.5/91.3  & \textbf{99.6}/\textbf{95.8} \\
 screw    & \textbf{99.7}/\textbf{72.9} & 99.1/62.0 &  98.7/61.4  &  99.1/51.8  &  98.7/34.2 & 98.9/31.7 &  66.5/21.1  &  51.0/5.1  &  97.5/44.9  &  98.1/57.1  \\
 tile   & 99.4/96.4 &  99.2/96.3  & 98.6/92.6 &  97.4/78.2  & 94.9/56.0 &98.3/80.6 & 88.7/63.9  & 91.0/54.4 & 99.6/96.5 &  \textbf{99.7}/\textbf{97.1}  \\
 toothbrush   &  97.3/49.2  &  97.5/38.9  &  98.4/61.7  & 99.0/63.1 &   97.6/37.1  & 99.4/64.2 & 96.3/52.4  &  74.5/4.8  & \textbf{99.6}/\textbf{78.1} &  98.2/68.3  \\
 transistor  &  92.2/56.0  &  85.3/36.5  &  98.6/82.9  &  \textbf{99.6}/50.3 & 91.8/66.7 & 92.5/61.2 & 55.2/4.4  & 79.3/11.2 & 98.4/85.6 & 98.0/\textbf{86.7} \\
 wood   & 97.6/81.6 & 97.2/77.1 & 97.6/25.6 &  99.3/39.1  &  95.7/54.3  & 98.7/77.5 & 93.1/47.9 & 82.9/21.0 &  97.8/82.6  & \textbf{99.4}/\textbf{91.6} \\
 zipper   & 98.6/73.6 &  98.1/78.2  & 95.9/53.9 &  \textbf{99.7}/52.7  &  98.5/63.1  & 98.4/64.2 & 92.4/53.1  & 96.8/42.3 & 98.8/77.6 & 99.6/\textbf{90.7} \\
\hline 
Average   & 97.7/69.0 &  96.2/65.5  &  98.3/66.3  &  98.3/57.8  &  97.1/56.6 & 97.5/62.6  & 86.4/49.3 &  84.8/25.7  & 99.0/78.6 & \textbf{99.1}/\textbf{84.5} \\
\bottomrule
\end{tabular}
\vspace{-0.2cm}
\caption{\textbf{Quantitative result for pixel-level anomaly localization (AUROC/AP)}. We employ a simple U-Net trained on our generated dataset and the existing anomaly detection methods with their official codes or pre-trained models.}
\label{detection models}
\vspace{-0.5cm}
\end{table*}

Moreover, we present the anomaly images generated by several prominent anomaly image generation models in Fig. \ref{fig:generations}. 
It is observed that the images generated by Defect-GAN often do not have anomalies, DFMGAN tends to introduce noise in the image (bottle, capsule), and the anomaly image-mask pairs lack proper alignment (grid). 
The state-of-the-art model AnomalyDiffusion sometimes generates unreasonable masks outside the objects (grid, screw), and the generated anomaly does not fit the original image enough to make the anomaly look unreal (toothbrush). 
In contrast, our proposed model demonstrates the ability to generate highly realistic and valid anomalies, even excelling in handling smaller-scale anomalies such as grid-bent.
More results are presented in the supplementary material. 

\noindent\textbf{Anomaly Generation for Anomaly Detection and Localization.}
We evaluate the effectiveness of our approach by comparing it with existing methods for anomaly generation in downstream anomaly detection and localization.
We calculate pixel-level dimensions and image-level AUROC, AP, and F1-max for each category. The results are presented in Tab. \ref{localization}, and because the image-level results of  AUROC, AP, and F1-max are relatively similar, as well as the space limitation, we only show the results of the most representative AP, and the other results of AUROC and F1-max are shown in the supplementary material. It can be seen that the U-net trained in our generated data reaches the highest AP of \textbf{84.5\%} and F1-max of \textbf{78.8\%}, surpassing the second-ranked AnomalyDiffusion by a margin of \textbf{3.1\%(AP)}.

\noindent\textbf{Anomaly Generation for Anomaly Classification.}
To further assess the quality of the anomaly images generated by our model, we utilize them to train a downstream anomaly classification model. Following the experimental setup of DFMGAN \cite{duan2023DFMGAN}, we employ ResNet-18 to train on the generated dataset. We then evaluate the classification accuracy on our test dataset as shown in Tab. \ref{classification}. 
As evident from the data, our model demonstrates significantly higher accuracy across the majority of categories compared to other models, with an average accuracy improvement of \textbf{13.58\%}, indicating the anomaly data we generate is more realistic.


\begin{figure}[!t]
\centering
\includegraphics[width=1\linewidth]{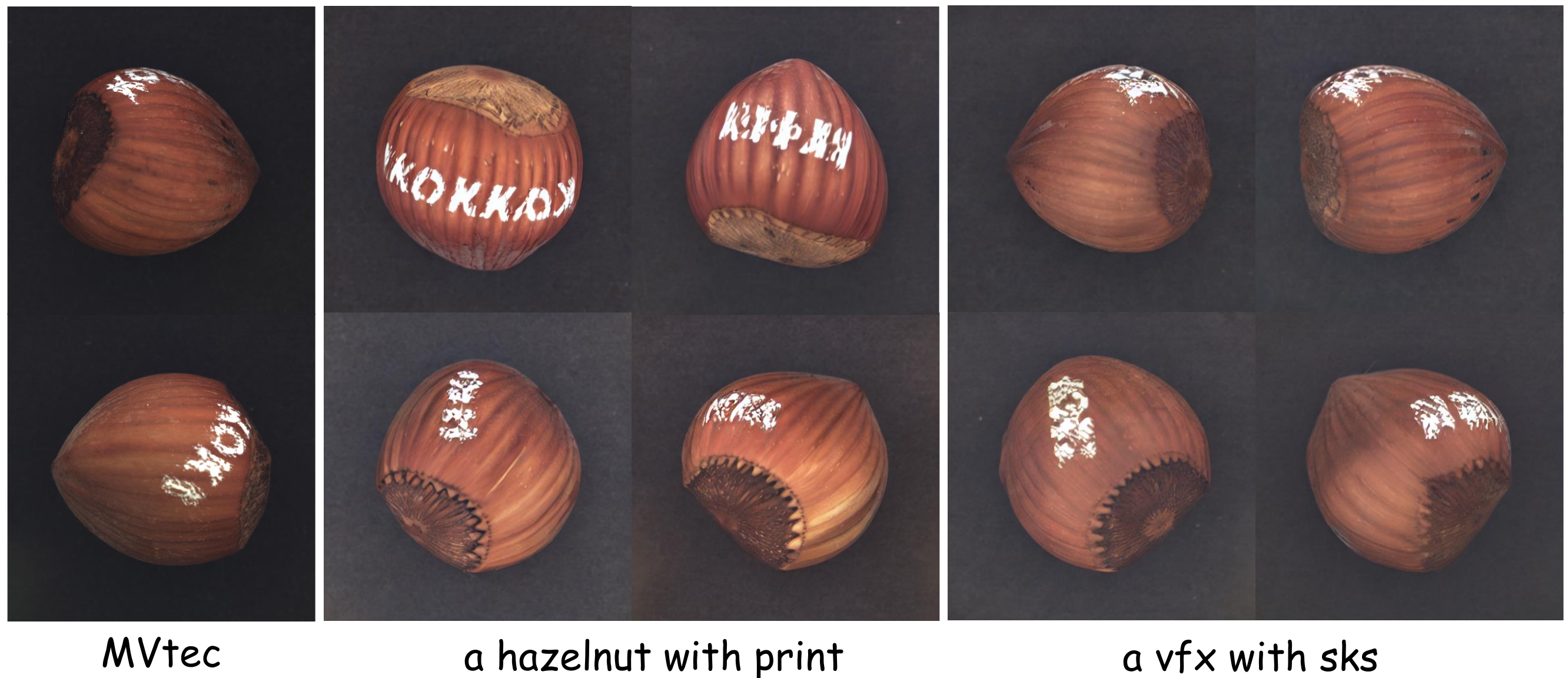}
\vspace{-0.5cm}
\caption{Comparison of generation results for different prompts on the hazelnut-print category.}
\label{fig:ablation_prompt}
\vspace{-0.5cm}
\end{figure}

\subsection{Comparison with Anomaly Detection Models}
To further validate the effectiveness of our model, we compare it with state-of-the-art anomaly detection methods DRAEM\cite{zavrtanik2021draem}, SSPCAB \cite{ristea2022sspcab}, CFA \cite{lee2022cfa}, RD4AD \cite{deng2022rd4ad}, PatchCore \cite{roth2022patchcore}, MuSc \cite{li2024musc}, DevNet \cite{pang2021devnet}, DRA \cite{ding2022dra} and PRN \cite{zhang2023prn}. We utilize their official codes or pre-trained models and evaluate them on the same testing dataset (2/3 anomaly data in the test of  MVTec). As there is no available open-source code for PRN, we rely on the results provided in its research paper. The methods are evaluated based on their pixel-level AUROC and AP scores, as demonstrated in Tab. \ref{detection models}.
It is clearly discernible that upon using the anomaly data generated by our method, even if only the simplest U-Net model is utilized, the result shows a significant increase of \textbf{5.9\%} over that of the traditional method.


\subsection{Ablation Study}
Fig. \ref{fig:ablation_prompt} presents the results of using different prompts in our model. It is evident that using the prompt ``a vfx with sks''  yields superior performance in terms of object shape, color, and other aspects compared to using the category name directly. The reason is that using real object names and defect category names as prompt introduces a significant amount of prior knowledge. However, the prior knowledge from the natural world may differ greatly from the specific objects and defects in the specific dataset. More quantitative results can be found in the supplementary material. 

\begin{table}[t]
\setlength{\tabcolsep}{0.6mm}
\centering
\small
\begin{tabular}{c|ccc|ccc}
\toprule 
\multirow{2}{*}{ Category } & \multicolumn{3}{c|}{Without BCM} & \multicolumn{3}{c}{With BCM} \\
 & AUROC &  AP &  $F_{1}$-max & AUROC &  AP &  $F_{1}$-max \\
\midrule
bottle & 98.4 & 88.8 & 77.1 &\textbf{ 99.5} & \textbf{93.4} & \textbf{85.7}  \\
pill &  98.4 & 86.9 & 78.2 & \textbf{99.6} & \textbf{95.8} &  \textbf{89.2}  \\
toothbrush & 97.2 & 62.7 & 64.0 & \textbf{98.2} & \textbf{68.3} & \textbf{68.6}  \\
Average & 98.8 & 83.0 & 78.1 & \textbf{99.1}& \textbf{84.5} & \textbf{78.8}\\
\bottomrule
\end{tabular}
\vspace{-0.2cm}
\caption{Comparison on pixel-level anomaly localization on part of categories in MVTec between with/without the Background Compensation Module. Average is the mean of all categories.}
\label{table:bcm}
\vspace{-0.6cm}
\end{table}

Furthermore, we assess the effectiveness of the Background Compensation Module (BCM), which exhibits more noticeable improvements in the categories of bottle, pill, and toothbrush. The visualization results are depicted in Fig. 4 and the pixel-level localization results are presented in Tab. 5. All the qualitative image generation results and quantitative metrics are improved significantly by using BCM, demonstrating BCM is critical for certain categories.


\section{Conclusion}
In this paper, we present a novel approach, DualAnoDiff, for generating anomalous image-mask pairs. Our method employs a parallel dual-diffusion to simultaneously generate the anomaly image and the corresponding anomaly part. This ensures a high level of alignment between the generated anomaly image-mask pair and the realism of the anomaly image. Additionally, to address challenging cases, we introduce a background compensation module that efficiently enhances the model's fitting capability.
Extensive experiments demonstrate its superior performance compared to existing anomaly generation methods. The anomaly data generated by our model effectively enhances the performance of downstream anomaly inspection tasks. 
{
    \small
    \bibliographystyle{ieeenat_fullname}
    \bibliography{main}
}
\clearpage
\setcounter{page}{1}
\maketitlesupplementary
\section{Implementation Details}
\subsection{Training Details}
We train a set of model parameters for each anomaly type. The model requires 5,000 epochs for training, which takes approximately 4.5 hours on an NVIDIA V100 32GB GPU. With BCM, the training step requires only 2,000 epochs. The batch size is set to 4, the learning rate is 0.000005, and the rank of LoRA is 32. 

During training, 
we utilize random flipping for data augmentation.
For the ``Background Compensation Module'', we have applied it to all categories that involve backgrounds. Among them, categories bottle, pill and toothbrush have witnessed a highly significant improvement. The improvement in the other several categories is rather limited since they can already be generated with a high level of quality.

\subsection{Inference Details}
During inference without BCM, we only need to input a set of prompts: ``a vfx with sks'' and ``sks''. This process generates a set of anomaly images along with the corresponding anomaly part images. 
We generate 1000 image pairs with a resolution of 512×512 for each anomaly. Specifically, the num\_inference\_steps is set to 50, and the guidance\_scale is set to 2.5. Notably, it takes 15 seconds to generate each pair of images.

\subsection{Mask Generation}
We employ U$^2$-Net \cite{qin2020u2} to segment the anomaly part image and obtain the corresponding mask. Based on our observations, this mask is entirely accurate.

\section  {More Ablation Studies}
We present comprehensive pixel-level and image-level results for downstream anomaly detection in Tables \ref{a-1} and \ref{a-2}. The term ``dual-interrelated diffusion'' refers to the utilization of the dual-interrelated model framework, where the type name such as ``cable'' is employed as a prompt. The notation ``+ prompt'' indicates the replacement of the type name with ``vfx'' and ``sks''. Additionally, ``+BCM'' signifies the incorporation of the Background Compensation Module, which is specifically applied to the categories of bottle, grid, hazelnut, pill, and screw. It can be observed that the prompt we designed outperforms the use of category names, with the exception of the toothbrush category. However, the gap between the toothbrush category and the prompt can be effectively bridged by the BCM module.
\begin{table*}[t]
\small
\centering
\begin{tabular}{c|ccc|ccc|ccc}
\toprule 
\multirow{2}{*}{ Category } &\multicolumn{3}{c}{dual-interrelated diffusion} & \multicolumn{3}{c}{ 
$+$prompt} & \multicolumn{3}{c}{$+$prompt $+$BCM}  \\
  & AUC-P & AP-P &  $F_{1}$-P &  AUC-P &AP-P &$F_{1}$-P & AUC-P & AP-P &  $F_{1}$-P \\
\midrule
bottle & 96.4& 74.2& 69.7&  98.4&88.8&77.1& \textbf{99.5} & \textbf{93.4}& \textbf{85.7}\\
cable  & 95.7& 74.1& 68.8&  \textbf{97.5}&\textbf{82.6}&\textbf{76.9}& \textbf{97.5}& \textbf{82.6}& \textbf{76.9}\\
capsule  & 97.8& 54.8& 54.3&  \textbf{99.5} &\textbf{73.2} &\textbf{67.0} 
& \textbf{99.5} & \textbf{73.2} & \textbf{67.0} \\
carpet  & 99.4& 86.7& 77.9&  \textbf{99.4} &\textbf{89.1} &\textbf{80.2} 
& \textbf{99.4} & \textbf{89.1} & \textbf{80.2} \\
grid  & 95.8& 36.2& 39.8&  98.5 &57.2 &54.9 
& \textbf{98.5} & \textbf{57.2} & \textbf{54.9} \\
hazelnut  & 99.5& 94.8& 89.9&  \textbf{99.8} &96.5&91.5& \textbf{99.8} & \textbf{97.7} & \textbf{92.8} \\
leather  & 98.4& 79.1& 70.1&  \textbf{99.9} &\textbf{88.8} &\textbf{78.8} 
& \textbf{99.9} & \textbf{88.8} & \textbf{78.8} \\
metal\_nut  & 98.8& 94.4& 89.1&  \textbf{99.6}&\textbf{98.0} &\textbf{93.0} 
& \textbf{99.6}& \textbf{98.0 }& \textbf{93.0} \\
pill  & 89.6& 38.1& 31.2&  98.4&86.9&78.2& \textbf{99.6}& \textbf{95.8}& \textbf{89.2}\\
screw  & 97.7& 48.9& 47.9&  97.7&55.15&72.8& \textbf{98.1} & \textbf{57.1} & \textbf{56.1} \\
tile  & 99.1& 91.0& 80.8&  \textbf{99.7} &\textbf{97.1} &\textbf{91.0}& \textbf{99.7} & \textbf{97.1} & \textbf{91.0}\\
toothbrush  & 98.2& 65.2& 67.1&  97.2&62.7&64.0& \textbf{98.2}& \textbf{68.3} & \textbf{68.6} \\
transistor  & 94.9& 78.2& 73.1&  \textbf{98.0}&\textbf{86.7}&\textbf{79.6}& \textbf{98.0}& \textbf{86.7}& \textbf{79.6}\\
wood  & 98.6& 87.3& 75.9&  \textbf{99.4} &\textbf{91.6} &\textbf{83.8} 
& \textbf{99.4} & \textbf{91.6} & \textbf{83.8} \\
zipper  & 98.4& 82.2& 72.5&  \textbf{99.6} &\textbf{90.7} &\textbf{82.7} 
& \textbf{99.6} & \textbf{90.7} & \textbf{82.7} \\
\hline
Average  & 97.22& 72.35& 67.21&  98.8&83.0&78.1& \textbf{99.1} & \textbf{84.5} & \textbf{78.8} \\
\bottomrule
\end{tabular}
\caption{Ablaiton Study: comparison on pixel-level anomaly localization
on the MVTec dataset by training a U-Net on our model's generated data using different settings.}
\label{a-1}
\vspace{-0.3cm}
\end{table*}

\begin{table*}[t]
\small
\centering
\begin{tabular}{c|ccc|ccc|ccc}
\toprule 
\multirow{2}{*}{ Category } &\multicolumn{3}{c}{dual-interrelated diffusion} & \multicolumn{3}{c}{$+$prompt} & \multicolumn{3}{c}{$+$prompt $+$BCM} \\
  & AUC-P & AP-P &  $F_{1}$-P &  AUC-I&AP-I  &F$_{1}$-I& AUC-P & AP-P &  $F_{1}$-P \\
\midrule
bottle & 98.0& 99.2& 96.4&  98.7&98.0&98.9& \textbf{100}& \textbf{100}  & \textbf{100}
\\
cable  & 92.3& 94.5& 85.1
&  \textbf{97.7}&\textbf{98.3}&\textbf{94.2}
& \textbf{97.7}&\textbf{ 98.3}& \textbf{94.2}
\\
capsule  & 81.9& 93.5& 88.9
&  \textbf{97.6}&\textbf{99.2}&\textbf{95.8}
& \textbf{97.6}& \textbf{99.2}& \textbf{95.8}
\\
carpet  & 96.7& 98.8& 95.7
&  \textbf{99.8}&\textbf{99.9}  &\textbf{99.1}
& \textbf{99.8}& \textbf{99.9}  & \textbf{99.1}
\\
grid  & 97.2& 98.6& 95.0
&  \textbf{99.5}&\textbf{99.7}&\textbf{97.6}
& \textbf{99.5}& \textbf{99.7}& \textbf{97.6}
\\
hazelnut  & \textbf{100}& \textbf{100}& \textbf{100}
&  \textbf{100}&\textbf{100}&\textbf{100}& \textbf{100}& \textbf{100}  & \textbf{100}
\\
leather  & \textbf{100}& \textbf{100}  & \textbf{100}
&  \textbf{100}&\textbf{100}  &\textbf{100}
& \textbf{100}& \textbf{100}  & \textbf{100}
\\
metal\_nut  & 97.7& 99.3& 97.6&  \textbf{99.7}&\textbf{99.9}&\textbf{99.2}
& \textbf{99.7}&\textbf{ 99.9}& \textbf{99.2}
\\
pill  & 87.1& 96.3& 91.2
&  92.0&97.8&93.6& \textbf{95.8}& \textbf{99.0}& \textbf{95.8}
\\
screw  & 83.5& 90.1& 84.1
&  86.6&94.2&86.1& \textbf{87.8}& \textbf{95.0 } & \textbf{87.2}
\\
tile  & \textbf{100}& \textbf{100}  & \textbf{100}
&  \textbf{100}&\textbf{100}  &\textbf{100}
& \textbf{100}& \textbf{100}  & \textbf{100}
\\
toothbrush  & 97.9& 98.8& 94.7
&  97.6&98.5&93.9& \textbf{99.5}& \textbf{99.7}  & \textbf{97.5}
\\
transistor  & 92.8& 92.3& 89.4
&  \textbf{95.1}&\textbf{93.7}&\textbf{90.1}
& \textbf{95.1}& \textbf{93.7}& \textbf{90.1}
\\
wood  & 99.3& 99.7& 97.6
&  \textbf{100}&\textbf{99.9}&\textbf{100}
& \textbf{100}& \textbf{99.9}& \textbf{100}
\\
zipper  & \textbf{100}& \textbf{100}  & \textbf{100}&  \textbf{100}&\textbf{100}  &\textbf{100}& \textbf{100}& \textbf{100}  & \textbf{100}\\
\hline
Average  & 94.9& 97.4& 94.38&  97.6&98.6&96.5&\textbf{ 99.8}&\textbf{ 98.9}& \textbf{99.8}\\
\bottomrule
\end{tabular}
\caption{Ablaiton Study: comparison on image-level anomaly localization
on the MVTec dataset by training a U-Net on our model's generated data using different settings.}
\label{a-2}
\vspace{-0.3cm}
\end{table*}

\section{More Qualitative Experiments}
We conducted a comprehensive comparison between our generated results and those of existing anomaly image generation methods, with the results presented in Fig. \ref{cmp}. It is evident that the diversity of anomalies generated by Crop\&Paste \cite{lin2021croppaste} is limited. The results from DiffAug \cite{zhao2020diffaug} exhibit overfitting. The generated outcomes from CDC \cite{ojha2021few} lack realism, often resulting in distortion, deformation, and other artifacts. SDGAN \cite{niu2020sdgan} and Defect-GAN \cite{zhang2021defectgan} fail to generate masks corresponding to the anomalies, and the authenticity of the generated images is also limited. The masks produced by DFMGAN \cite{duan2023DFMGAN} are not sufficiently aligned, often resulting in the generation of spots or noise. The currently best-performing method, AnomalyDiffusion \cite{hu2024anomalydiffusion}, solely focuses on learning the anomaly part. Consequently, the generated anomaly data fails to integrate smoothly with the original image. And, this sometimes leads to the situation where anomalies manifest against the backdrop of the image.
In contrast, our method not only generates highly realistic and diverse anomaly data but also produces highly aligned corresponding masks.

Among all these methods, DFMGAN and AnomalyDiffusion are currently the two best performers, so we conducted a more detailed visualization comparison of our results with these two methods. Additional visualizations are presented in Fig. \ref{cmp-1}-\ref{cmp-6}. The left side shows two examples from the training data, while the right side displays the generated image pairs.

\begin{figure*}[!ht]
\centering
\includegraphics[width=1\textwidth]{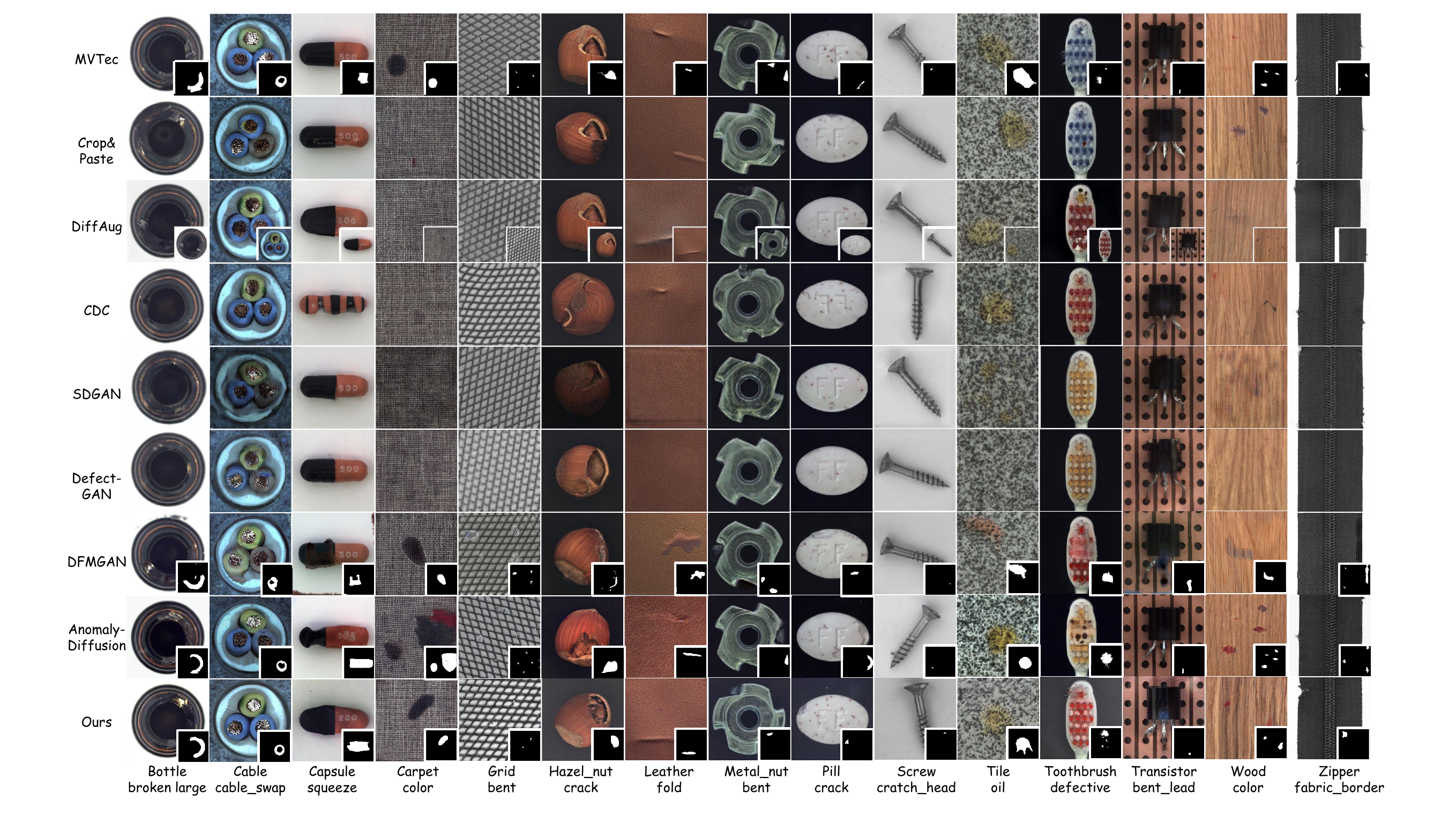} 
\caption{Comparison on the generation results on MVTec.}
\label{cmp}
\end{figure*}

\begin{figure*}[!ht]
\centering
\includegraphics[width=1\textwidth]{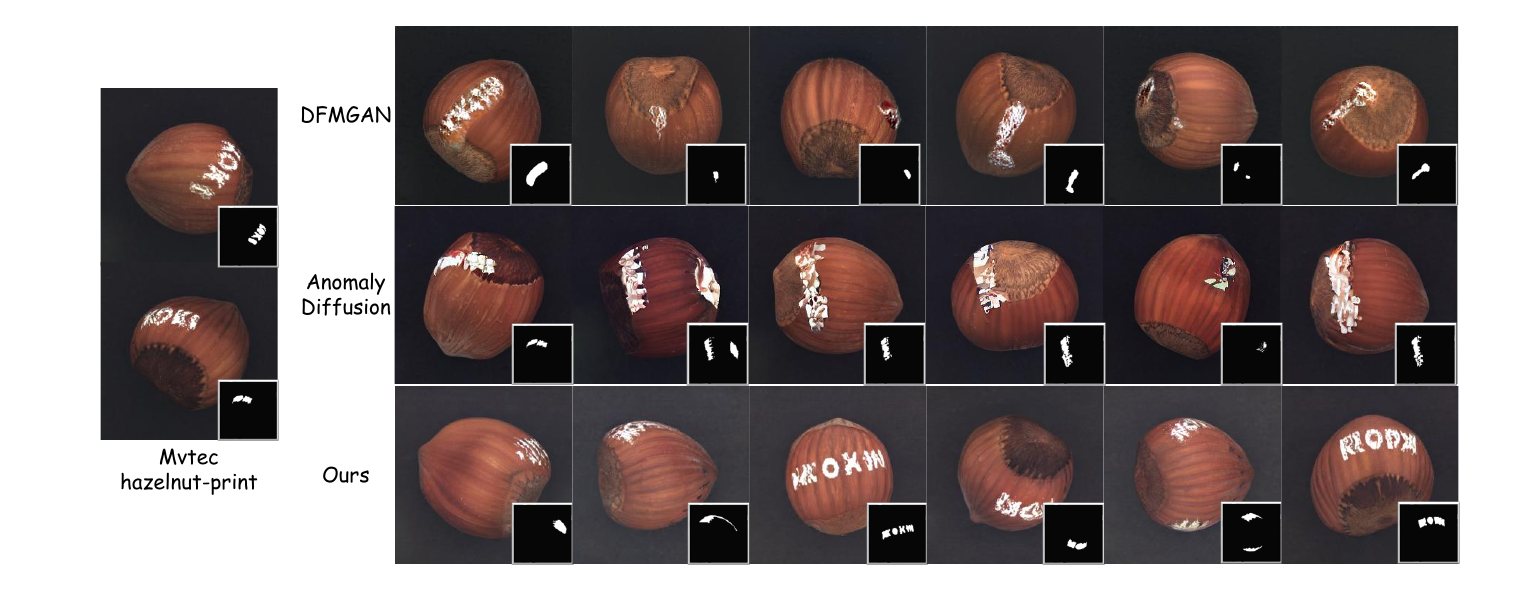} 
\caption{Comparison on the type of hazelnut-print. DFMGAN and AnomalyDiffusion struggle to generate realistic anomalies, particularly in the print class, where the anomalous regions consist of strings of letters. In contrast, our method successfully generates both the shape of the letters and the corresponding mask that aligns with their contours. }
\label{cmp-1}
\end{figure*}

\begin{figure*}[!ht]
\centering
\includegraphics[width=1\textwidth]{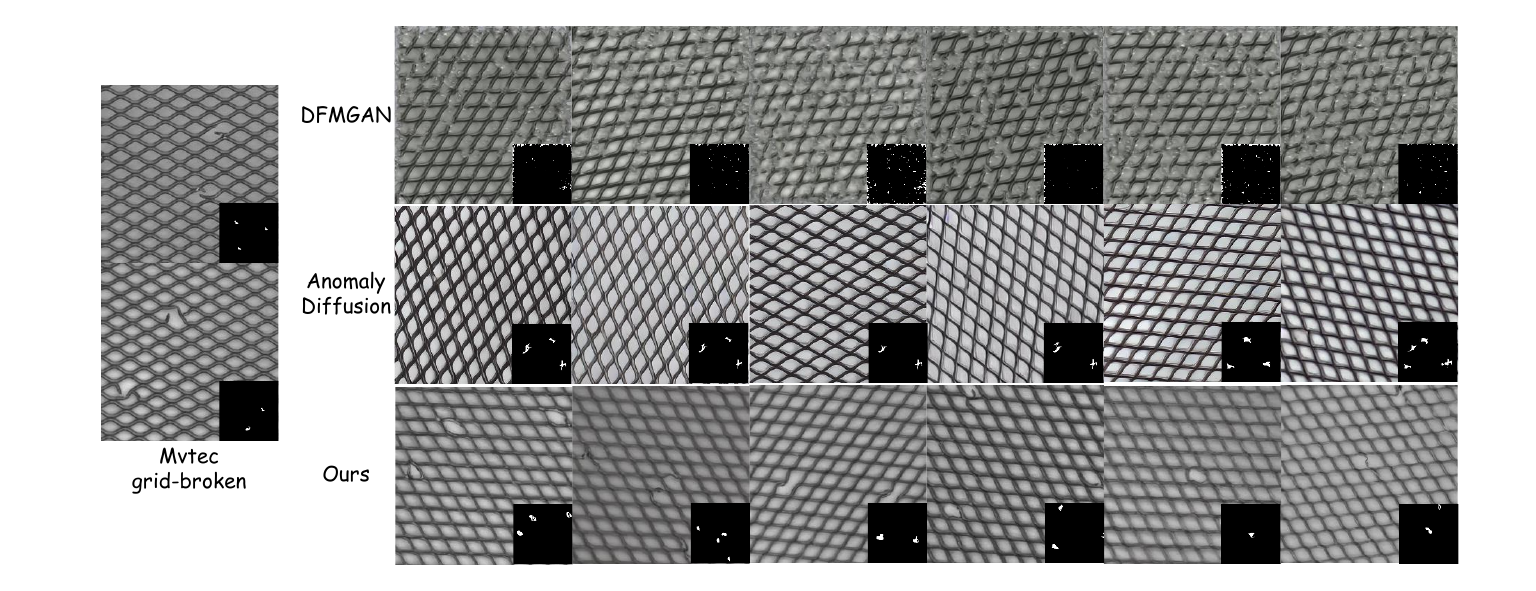} 
\caption{Comparison on the type of grid-broken. For this specific type of small, structure-related anomaly, the images generated by DFMGAN are of poor quality, and AnomalyDiffusion fails to produce any anomalies. In contrast, our method generates highly realistic and effective anomaly images.}
\label{cmp-2}
\end{figure*}

\begin{figure*}[!ht]
\centering
\includegraphics[width=1\textwidth]{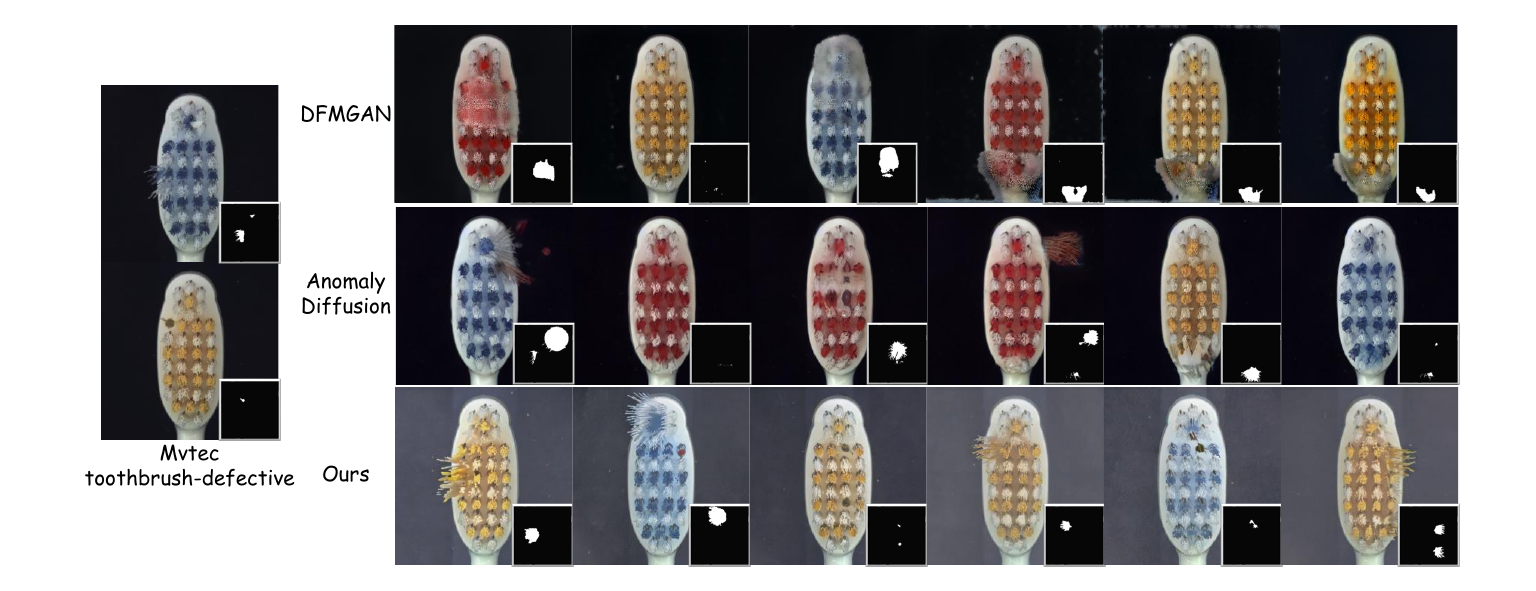} 
\caption{Comparison on the type of toothbrush-defective. The anomalies generated by DFMGAN lack realism, while those produced by AnomalyDiffusion are detached from the main object, Additionally, the generated anomalies, such as holes and bristles of toothbrushes, are mixed. In contrast, although there are some differences in background color, the generated anomalies by our model are fully consistent with real-world scenarios. Furthermore, the background issues do not impact the effectiveness of anomaly detection in downstream tasks.}
\label{cmp-3}
\end{figure*}

\begin{figure*}[!ht]
\centering
\includegraphics[width=1\textwidth]{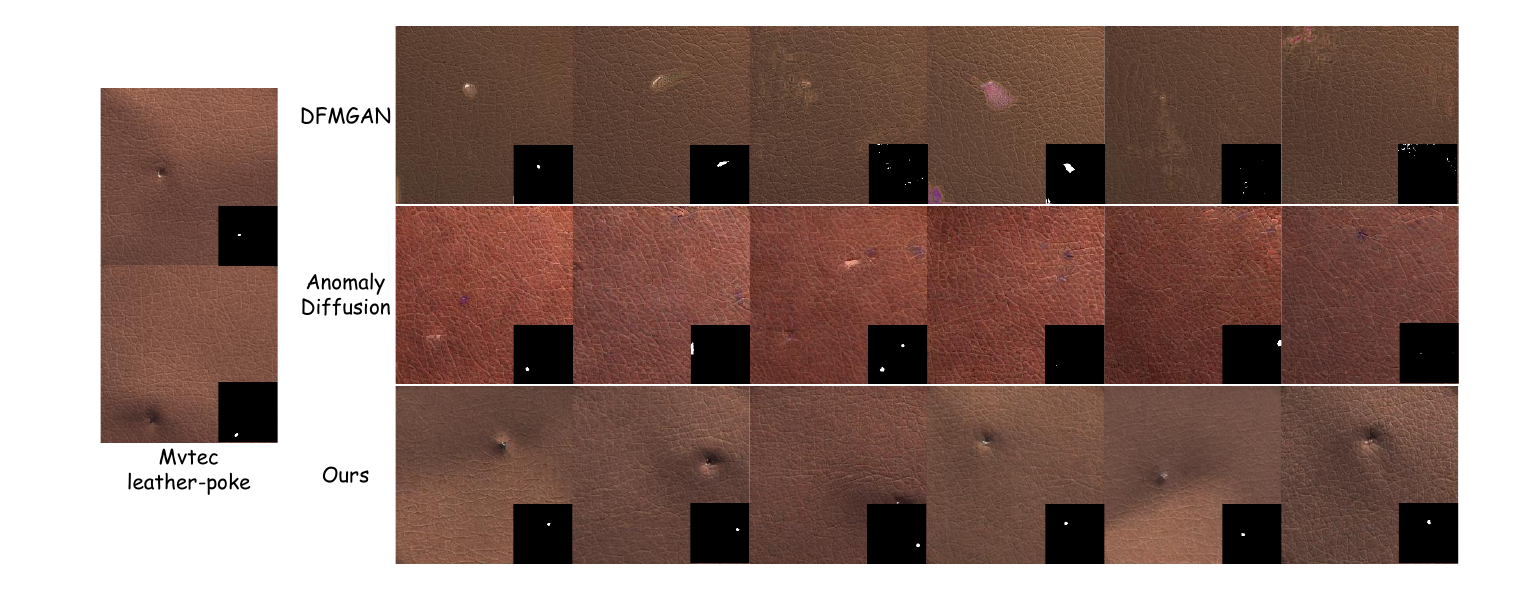} 
\caption{Comparison on the type of leather-poke. The anomalies generated by AnomalyDiffusion are slightly better than those produced by DFMGAN, however, there is a noticeable color difference in the leather. In contrast, our method achieves good results in both aspects.}
\label{cmp-4}
\end{figure*}

\begin{figure*}[!ht]
\centering
\includegraphics[width=1\textwidth]{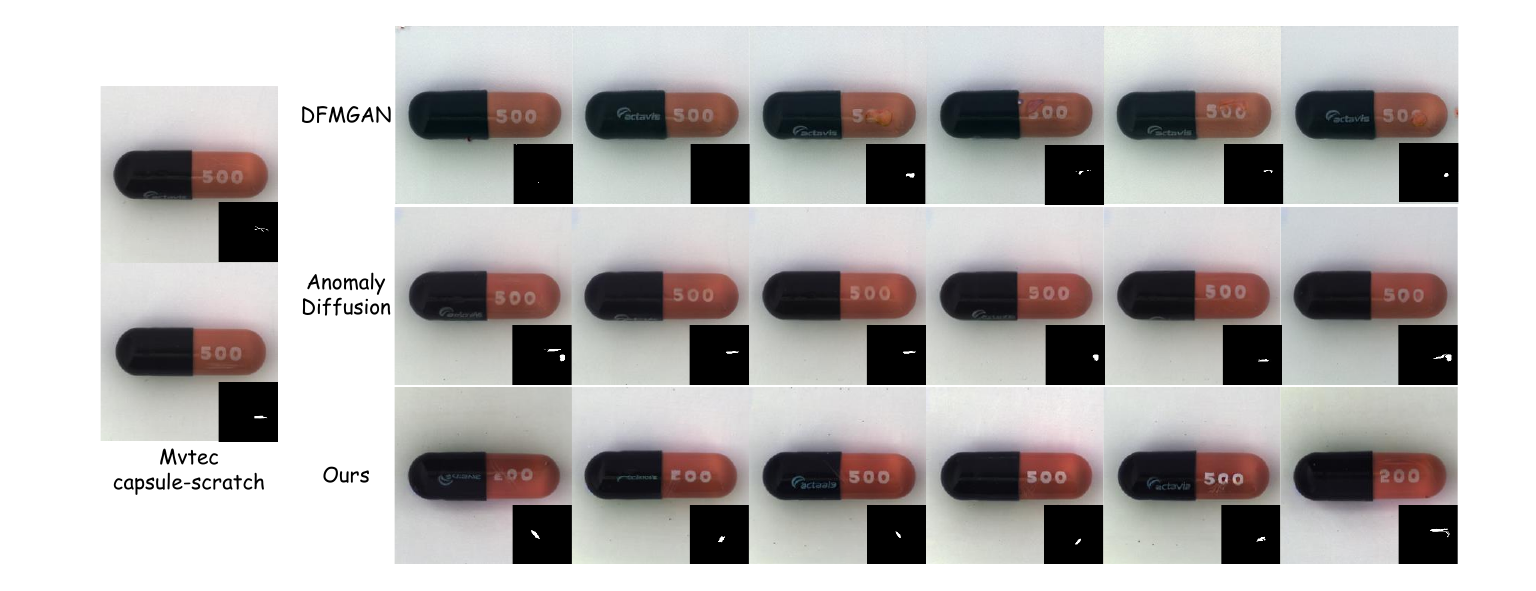} 
\caption{Comparison on the type of capsule-scratch. For scratches, a relatively minor type of anomaly, neither DFMGAN nor AnomalyDiffusion can generate effective results. In contrast, our method not only produces realistic anomalies but also demonstrates a good variety.}
\label{cmp-5}
\end{figure*}

\begin{figure*}[!ht]
\centering
\includegraphics[width=1\textwidth]{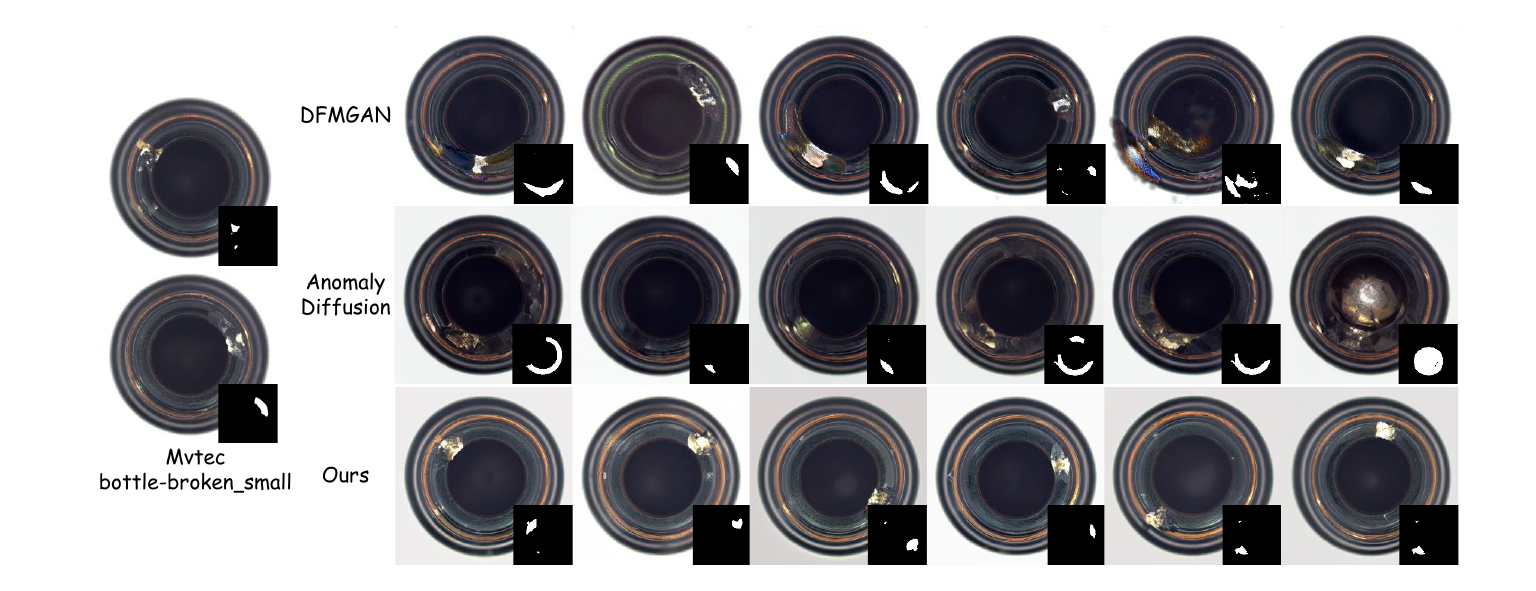} 
\caption{Comparison on the type of bottle-broken\_small. This type of anomaly refers to a small blemish around the edge of a bottle, while broken\_large indicates a larger blemish in the same area. The quality of the image generated by DFMGAN is limited, and the mask are not properly aligned. 
while the abnormal position generated by AnomalyDiffusion sometimes is not correct, and the shape does not belong to the type of broken\_small, but more like broken\_large. Our method, however, achieves good results in both position and shape.
}
\label{cmp-6}
\end{figure*}

\begin{figure*}[!ht]
\centering
\includegraphics[width=1\textwidth]{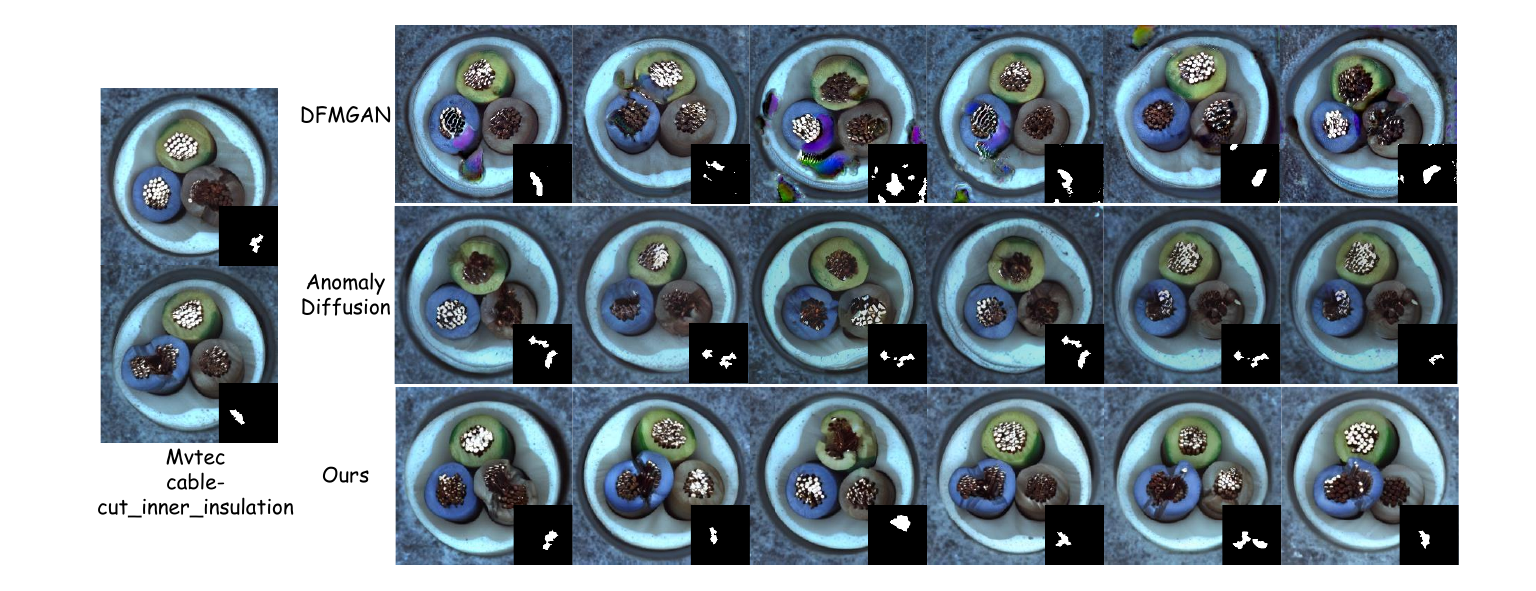} 
\caption{Comparison on the type of cable-cut\_inner\_insulation. It is evident that neither DFMGAN nor AnomalyDiffusion can generate realistic anomalies, and the overall quality of the images produced by DFMGAN is subpar. In contrast, our method successfully generates realistic and diverse abnormal data.
}
\label{cmp-6}
\end{figure*}

\section{Quantitative Experiments Setting}
\subsection{Generated Data}
In all comparison methods, 1000 sets of data are generated for each subclass for downstream detection tasks.

\subsection{Metrics}
This section provides supplementary information on the rationale for using these indicators and their definitions.\\
\textbf{For Generation.} 
General image generation tasks typically use \textbf{Fréchet Inception Distance (FID)} \cite{heusel2017fid} to evaluate the difference between the generated data and the real data distribution. However, FID is not reliable in cases of limited anomalous data, as it tends to produce higher scores for overfitted models. Therefore, we utilize the \textbf{Inception Score (IS)} \cite{salimans2016improved} as our evaluation metric. The IS does not require training data and quantifies the quality and diversity of the generated images by calculating the negative exponent of the Kullback-Leibler (KL) divergence between the edge distribution of the generated images and the conditional distribution of the class labels predicted by the Inception model. A higher IS score indicates better quality and diversity in the generated images.

In addition, we use \textbf{Intra-cluster Pairwise LPIPS Distance (IC-LPIPS)} \cite{ojha2021few} to measure the diversity of the generated data. This method clusters the images into k groups based on the LPIPS distance to k target samples and then computes the average mean LPIPS distances to the corresponding target samples within each cluster. Higher IC-LPIPS scores indicate better diversity.\\
\textbf{For Anomaly Inspection.}
We use the \textbf{Area Under the Receiver Operating Characteristic (AUROC)},  \textbf{Average Precision (AP)}, and \textbf{F$_{1}$-max} to measure the performance of the inspection following the general anomaly inspection task. 

\subsection{Anomaly Inspection Detail}
In the downstream task of anomaly detection, we employ a simple U-Net \cite{ronneberger2015unet} architecture. To mitigate the effects of randomness, we train the model three times and select the best result as the final outcome.


\end{document}